\documentclass[10pt,journal,compsoc]{IEEEtran}
\usepackage{amsmath,amsfonts}
\usepackage{algorithmic}
\usepackage{algorithm}
\usepackage{array}
\usepackage[caption=false,font=footnotesize,labelfont=sf,textfont=sf]{subfig}
\usepackage{textcomp}
\usepackage{stfloats}
\usepackage{url}
\usepackage{verbatim}
\usepackage{graphicx}
\usepackage{cite}
\usepackage{multirow}
\usepackage{pifont}
\usepackage{booktabs}
\usepackage{tabularx}
\usepackage[table, dvipsnames]{xcolor}

\usepackage[switch]{lineno}

\usepackage[dvipsnames]{xcolor}
\usepackage{xspace}
\usepackage{hyperref}

\usepackage{booktabs}
\usepackage{graphicx}

\usepackage{pgfplots}
\pgfplotsset{compat=1.17} % Set version compatibility

\usepackage[flushleft]{threeparttable}

\usepackage{soulpos}
\usepackage{comment}

\DeclareMathOperator*{\mean}{\mathtt{mean}}

\newcommand{\quotes}[1]{``#1''}

\newcommand{\ours}{Hier-EgoPack\xspace}

\newcommand{\ourscvpr}{EgoPack\xspace}
\newcommand{\egofourd}{Ego4D\xspace}

\usepackage{xspace}
\newcommand*{\eg}{e.g.\@\xspace}
\newcommand*{\ie}{i.e.\@\xspace}

\definecolor{up_color}{RGB}{204,51,0}
\definecolor{down_color}{RGB}{202,195,121}

\begin{document}
\linenumbers
\renewcommand\makeLineNumber{}

\title{Hier-EgoPack: Hierarchical Egocentric Video Understanding with Diverse Task Perspectives}

\author{Simone Alberto Peirone, Francesca Pistilli, Antonio Alliegro, Tatiana Tommasi and Giuseppe Averta
        % <-this % stops a space
\thanks{The authors are with the Department of
Control and Computer Engineering, at Politecnico di Torino, Italy. Email: \texttt{firstname.lastname@polito.it}.}}

\markboth{Journal of \LaTeX\ Class Files,~Vol.~14, No.~8, August~2021}%
{Peirone \MakeLowercase{\textit{et al.}}: Hier-EgoPack: Hierarchical Egocentric Video Understanding with Diverse Task Perspectives}

\IEEEpubid{0000--0000/00\$00.00~\copyright~2021 IEEE}

\IEEEtitleabstractindextext{%
\begin{abstract}
    Our comprehension of video streams depicting human activities is naturally multifaceted: in just a few moments, we can grasp what is happening, identify the relevance and interactions of objects in the scene, and forecast what will happen soon, everything all at once. To endow autonomous systems with such a holistic perception, learning how to correlate concepts, abstract knowledge across diverse tasks, and leverage tasks synergies when learning novel skills is essential.
    A significant step in this direction is \ourscvpr, a unified framework for understanding human activities across diverse tasks with minimal overhead. \ourscvpr promotes information sharing and collaboration among downstream tasks, essential for efficiently learning new skills.
    In this paper, we introduce \ours, which advances \ourscvpr by enabling reasoning also across diverse temporal granularities, which expands its applicability to a broader range of downstream tasks.
    To achieve this, we propose a novel hierarchical architecture for temporal reasoning equipped with a GNN layer specifically designed to tackle the challenges of multi-granularity reasoning effectively.
    We evaluate our approach on multiple \egofourd benchmarks involving both clip-level and frame-level reasoning, demonstrating how our hierarchical unified architecture effectively solves these diverse tasks simultaneously.
    Project page: \href{https://sapeirone.github.io/hier-egopack/}{sapeirone.github.io/hier-egopack}.
\end{abstract}
}

\maketitle

\section{Introduction}\label{sec:intro}
\IEEEPARstart{O}{ur}
daily activities are extremely complex and diverse, yet humans have the extraordinary ability to perceive, reason, and plan their actions almost entirely from visual inputs.
For instance, when observing someone at a kitchen counter with a pack of flour and a jug of water, we can infer they are kneading dough (\textit{reasoning about current activity}).
We might predict that their next step will involve mixing flour with water (\textit{reasoning about the future}) to obtain the dough (\textit{reasoning about implications}), maybe with the ultimate goal of preparing some bread (\textit{reasoning about long-range activities}).
Mastering such \quotes{skills} requires analyzing varying portions of the video and reasoning at different levels of granularity.
Long-term activities require analysis of a broader context over extended clips, while finer details, such as distinguishing when someone shifts from measuring flour to pouring water, call for reasoning at a frame level.
Such holistic reasoning, which is natural for humans, poses a significant challenge for artificial intelligence systems.
The difficulty lies in integrating various levels of reasoning, from low-level actions to high-level activity understanding, into a unified framework, while uncovering and leveraging the underlying semantic relationships between these skills to efficiently learn new ones by building on prior knowledge.

Current research trends in human activity understanding predominantly focus on creating several, hyper-specialized, models. This approach splits the understanding of human activities into distinct skills (\ie, tasks), for which each model is independently trained to rely only on \quotes{task-specific} clues for prediction~\cite{yan2022multiview,zhong2023anticipative,zhang2022actionformer}.
However, this approach overlooks that different tasks may share similar or complementary reasoning patterns, \ie, looking at the same video portion from different \textit{perspectives}.

To leverage the interplay between such different task perspectives, a first strategy might involve Multi-Task Learning (MTL), exploiting the intuition that knowledge sharing between tasks may be beneficial for each of them.
However, MTL suffers of some limitations~\cite{kokkinos2017ubernet}, mainly related to negative interferences between tasks, making it difficult to exploit task synergies effectively.
In addition, all task annotations must be available at training time, which hinders the extension of MTL models to novel tasks at a later point in time.

In the context of human behaviour understanding, usually inferred from videos collected in first person view, different tasks typically require closely related reasoning, resulting in a strong correlation between them. Consequently, studying and leveraging these inter-task synergies becomes particularly interesting.

\IEEEpubidadjcol
In this scenario, EgoT2 framework~\cite{egot2} represents an alternative solution to Multi-Task Learning, exploring how various egocentric video tasks can mutually benefit through the translation of task-specific cues across tasks.
However, although this approach fosters positive interactions between tasks, it has significant limitations: i) the primary task should be \quotes{known} at training time and present within the task-specific model collection, ii) it necessitates an extensive pretraining process and iii) it is inefficient as it relies on task-specific models instead of building transferable knowledge abstractions.

We argue that an important key to advancing the learning capabilities of intelligent systems and moving closer to more human-like reasoning lies not only in sharing information across tasks, but also in abstracting task-specific knowledge to make it reusable for learning novel tasks.
To enable this, we recently proposed \ourscvpr \cite{egopack}, a first effort in knowledge abstraction and sharing for egocentric videos understanding.
This method is able to exploit a set of known tasks (\emph{support tasks}), each one able to interpret an input stream according to its own task-perspective, to learn reusable knowledge abstractions that can aid in the learning of a \emph{novel task}. Such task-perspectives are encoded in the form of prototypes, collected in a single step from the pretraining of a multi-task network.
However, \ourscvpr implements limited temporal reasoning and, due to its flat architecture, cannot perform reasoning at different levels of granularity.
Notably, egocentric videos cover a wide range of tasks spanning diverse temporal scales, from sub-second actions to extended, long-range activities. While some tasks, such as action recognition and long-term anticipation, focus on fixed short segments, others, like temporal action localization, demand a more adaptive approach to deal with longer activities. As the temporal span of these tasks increases, developing a robust understanding of the sequential order of events, a concept known as \emph{sense of time}~\cite{bagad2023test}, becomes essential.

To address these challenges, we introduce \ours, an extension of \ourscvpr\cite{egopack}, specifically designed to maximize positive interaction across tasks with different temporal granularity, while still using a unified architecture and minimizing task-specific weights and tuning.
To achieve this, we present a hierarchical architecture that progressively learns more comprehensive representations of the input video, capturing both fine-grained details and broad contextual patterns.
A key aspect of this hierarchical design is effectively reasoning on temporal dependencies and consequentiality of actions, encompassing both past and future contexts.
To address this, we develop a novel GCN layer, hereinafter called Temporal Distance Gated Convolution (TDGC), specifically designed to encode these temporal relationships effectively.

We demonstrate the effectiveness and efficiency of our approach on \egofourd~\cite{ego4d}, a large-scale egocentric vision dataset.
To summarize, our main contributions are:
\begin{enumerate}
    \item We introduce a unified video understanding architecture to learn multiple egocentric vision tasks with different temporal granularity, while requiring minimal task-specific overhead;
    \item We present Temporal Distance Gated Convolution (TDGC), a novel GNN layer for egocentric vision tasks that require a strong \textit{sense of time};
    \item We extend \ourscvpr to the Moment Queries task, which involves the localization of activities that range from a few seconds to several minutes in duration;
    \item \ours achieves strong performance on five \egofourd~\cite{ego4d} benchmarks, using the same architecture and showing the importance of cross-task interaction.
\end{enumerate}

\section{Related works}\label{sec:related_works}

\subsection{Egocentric Vision}
Egocentric vision captures human activities from the privileged perspective of the camera wearer, allowing a unique point of view on their actions~\cite{betancourt2015evolution,plizzari2024outlook}.
Recently, the field has seen rapid development thanks to the release of several large-scale egocentric vision datasets~\cite{ek55,egtea,epic_tent,ek100,ego4d,sener2022assembly101}.
The rich annotations of these datasets~\cite{ek100,ego4d} allow to tackle a large number of tasks, including action recognition~\cite{nunez2022egocentric}, action anticipation~\cite{furnari2020rulstm,girdhar2021anticipative,zhong2023anticipative}, next active object prediction~\cite{furnari2017next}, action segmentation~\cite{zhang2022actionformer,huang2020improving}, episodic memory~\cite{ramakrishnan2023spotem} and long-range temporal reasoning tasks~\cite{goletto2025amego,mangalam2023egoschema,jia2022egotaskqa}.
Previous works in egocentric vision have focused on domain adaptation~\cite{Munro_2020_CVPR,yang2022interact,chen2019temporal,plizzari2023can,planamente2024relative}, multimodal learning~\cite{zehua2033human,gao2020listen, yang2022interact} and large-scale video-language pretraining~\cite{lin2022egocentric,pramanick2023egovlpv2,hiervl,zhao2023learning} to learn better representation for downstream tasks.

\subsection{Graph Neural Networks for vision tasks}
Traditional neural networks, including Convolutional Neural Networks (CNNs), have been widely used in computer vision, showing impressive performance on a variety of problems~\cite{li2021survey,khan2020survey,gu2018recent}.
However, these models often assume data lying on a regular domain, such as images that have a grid-like structure.
In recent years, the interest in developing methods able to provide a more general and powerful type of processing has been growing and particular attention has been given to learning methods on graphs.
Graph Neural Networks (GNNs) have the innate ability to effectively handle data that lie on irregular domains, such as 3D data~\cite{simonovsky2017dynamic,wang2019dynamic}, robotics~\cite{pistilli2023graph}, molecular chemistry~\cite{kearnes2016molecular}, and social or financial networks~\cite{fan2019graph}, and to model complex data relations~\cite{sanchez2020learning}.
Recently, transformer-based architectures had a great impact on vision applications.
Despite Transformers and GNNs share some similarities in their ability to handle various data types, they are fundamentally different in their core architectures and the specific ways they process data. GNNs can model the topology of a graph and the relations between nodes while also inheriting all the desirable properties of classic convolutions: locality, hierarchical structures and efficient weights reuse.
In video understanding, GNNs have been applied to action localization~\cite{huang2020improving,zeng2019graph,ghosh2020stacked,rashid2020action}, to build a knowledge graph from human actions~\cite{ghosh2020all}, to model human-object interactions~\cite{dessalene2020egocentric, dessalene2021forecasting} or to build a topological map of the environment~\cite{nagarajan2020ego}.

\subsection{Multi-Task Learning}
MTL~\cite{caruana1997multitask, zhang2021survey} tackles the problem of learning to solve multiple tasks simultaneously.
The development of this strategy is justified by the intuition that complex settings require solving multiple tasks, for instance autonomous driving~\cite{Huang_2023_ICCV}, robotics and natural language processing.
Furthermore, these networks can bring the theoretical advantage of sharing complementary information to improve performance.
Several works have been done in this direction~\cite{kokkinos2017ubernet, huang2020mutual, fifty2021efficiently, chen2022unified, Chen_2023_ICCV, shi2023deep, Huang_2023_ICCV, ci2023unihcp}, focusing on which parameters or tasks is better to share~\cite{kang2011learning, guo2020learning, standley2020tasks, sun2020adashare} and promoting synergies between tasks~\cite{kapidis2019multitask, wang2021interactive}.
Such methods encounter the problem of negative transfer~\cite{kokkinos2017ubernet} and sharing with unrelated tasks~\cite{guo2020learning, standley2020tasks} consequently suffering of task competition and not being able to benefit from information sharing between tasks. To overcome these limitations, several methods have been proposed to balance task-related losses~\cite{kendall2018multi, chen2018gradnorm, sinha2018gradient}, to dynamically prioritize tasks~\cite{guo2018dynamic}, to reduce gradient interference between tasks~\cite{gradient_surgery} or to exploit task interactions at multiple scales~\cite{vandenhende2020mti}.
Unfortunately, all these solutions require extensive task-specific tuning, and are not able to build an holistic perception across tasks.

Few works have explored MTL in egocentric vision~\cite{kapidis2019multitask,egot2,huang2020mutual,egopack}.
Among these, EgoT2~\cite{egot2} is the first to investigate semantic affinities among high-level egocentric vision tasks and learns how to translate the contributions of different task-specific models to support the learning of a primary task.

\ourscvpr~\cite{egopack} stands as a fundamentally different paradigm with respect to traditional MTL approaches by building a backpack of \emph{task perspectives} to leverage when learning a novel task.
This work further enhances the uniqueness of \ourscvpr paradigm by extending the support to tasks with diverse temporal granularities. To do this, we introduce a novel graph-based architecture that foster hierarchical temporal reasoning.

\section{Method}\label{sec:method}

\begin{figure}[t]
    \centering
    \includegraphics[trim=0 0.25cm 0 0,width=.95\columnwidth]{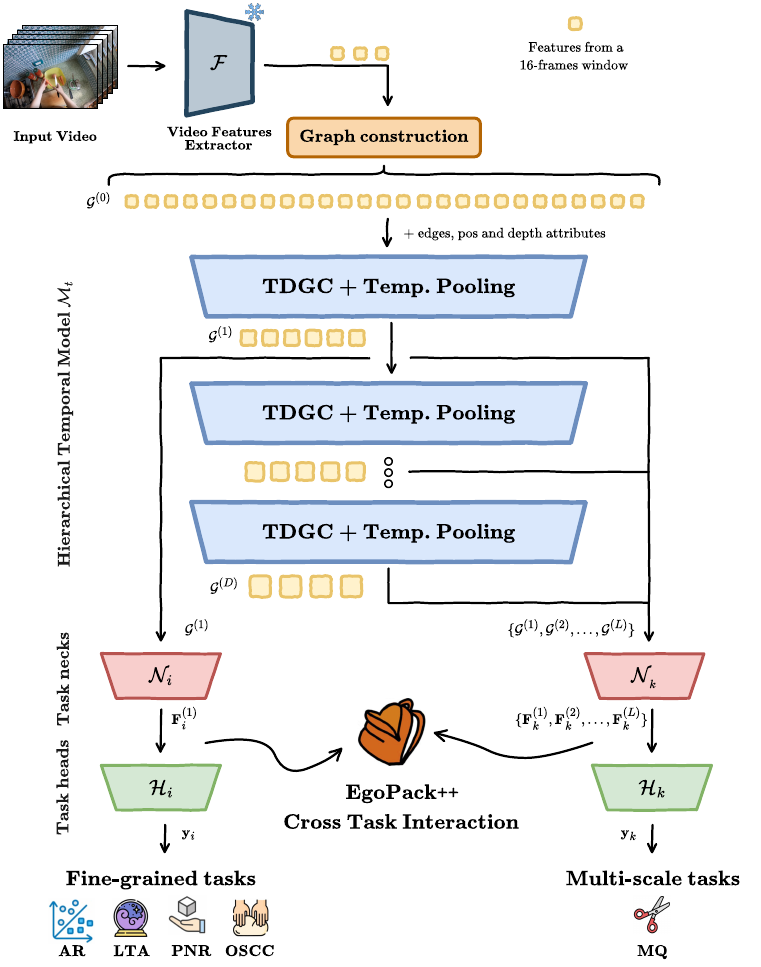}
    \caption{
        \textbf{Overview of the \ours architecture.}
        First, the video is converted into a graph representation $\mathcal{G}^{(0)}$ whose node embeddings are extracted using a frozen video features extractor.
        The graph is then processed by the \emph{hierarchical temporal backbone} $\mathcal{M}_t$, shared by all the tasks, to progressively learn higher level representations of the input video~$\{\mathcal{G}^{(1)}, \mathcal{G}^{(2)}, \dots, \mathcal{G}^{(L)}\}$.
        The node embeddings of these graphs are projected by the \emph{task-specific necks}~$\mathcal{N}_i$ in the features space of each task~$\mathcal{T}_i$ and to the corresponding output space with the \emph{task-specific heads}~$\mathcal{H}_i$.
    }\label{fig:architecture}
\end{figure}

We address a cross-task interaction setting, in which an egocentric vision model is trained to reuse previously acquired knowledge from a set of different tasks (\emph{support tasks}) to foster the learning process of any \emph{novel task}.
A formal definition of the proposed setting is presented in Sec.~\ref{sec:method_setting}.
This work introduces a unified temporal architecture to model tasks with different temporal granularity and strong \emph{sense of time}, \ie the ability to effectively reason on the order of the events in a video.
With this new architecture, we extend \ourscvpr to tasks that require long range temporal reasoning, \eg Temporal Action Localization.
We call this approach \ours, emphasizing its ability to learn hierarchical video representations that are well suited to various egocentric vision tasks.

\subsection{Setting: novel task learning}\label{sec:method_setting}
A task~$\mathcal{T}$ in egocentric vision is defined as a mapping between a video~$\mathcal{V}$ and an output space~$\mathcal{Y}$.
Classification tasks, such as Action Recognition, are defined as a mapping between a video segment~$v_i \in \mathcal{V}$ and the corresponding discrete label~$y_i \in \mathcal{Y}$. For these tasks, the start and end timestamps of the video segment $v_i$ are known.
Differently, the Temporal Action Localization (TAL) task processes the entire video~$\mathcal{V}$ and predicts a set of temporally grounded activities, each described by its start and end timestamps and the corresponding action label: $$\mathcal{T}: \mathcal{V} \to \{(t_i^s, t_i^e, y_i)\}_i.$$
We streamline the processing for different tasks by feeding our temporal backbone with untrimmed input videos and aligning the output to the downstream task at a later stage.
This alignment process is described more in depth in~Sec.~\ref{sec:method_ts}.

The cross-task interaction mechanism of \ours follows a two-stages training process.
First, a model $\mathcal{M}$ is trained on a set of $K$ tasks $\{\mathcal{T}_1,\dots,\mathcal{T}_K\}$, which we call \emph{support tasks}, in a Multi-Task Learning setting with hard-parameter sharing~\cite{ruder2017overview}.
The inclusion of multiple tasks in this phase encourages the model to learn more general and task-agnostic representations.
Then, the model is presented with a \emph{novel task} $\mathcal{T}_{K+1}$ to learn, without access to the supervision of the \emph{support tasks}.
In this scenario, the novel task can benefit from semantic affinities with the previously seen tasks.
For example, a model that has learned to detect object state changes may apply this knowledge for action recognition and vice-versa, as some actions produce object state changes, \eg \emph{cutting something}, while others do not, \eg \emph{moving an object}.
Our goal is to make these semantic affinities more explicit and exploitable, enabling the novel task to re-purpose these \emph{perspectives} from previous tasks to enhance performance,
a necessary step towards more holistic models that seamlessly share knowledge between~tasks.

\subsection{A unified architecture for Video Understanding}\label{sec:method_arch}
Egocentric vision tasks may provide complementary perspectives but also operate at different temporal granularities, from sub-second interactions to minutes-long activities.
To support all these tasks with a unified architecture, we need a model that can perform temporal reasoning hierarchically, progressively integrating fine-grained temporal representations into a broader and more comprehensive understanding.

Also, reasoning over long temporal horizons requires the ability to precisely ground and order past and future events.
The temporal backbone introduced in \ourscvpr~\cite{egopack} partially meets these constraints: while it supports multiple tasks with a shared architecture, it assumes similar temporal granularity across tasks and lacks a robust \emph{sense of time}, as detailed in Sec.~\ref{sec:exp_ablations}.
Indeed, the SAGE GNN convolutional operator used in \ourscvpr is invariant to permutations of the input nodes, and temporal ordering of the nodes is only provided by adding a positional encoding to the node embeddings.
This strategy is insufficient for tasks that require strong temporal reasoning, as we show in Sec.~\ref{sec:exp_ablations}.

We address these challenges by proposing a newly crafted hierarchical GNN-based architecture, specifically designed to support tasks with variable temporal resolution.
At the core we place a novel \textbf{Temporal Distance Gated Convolution (TDGC)} layer, able to explicitly encode past and future information, and a temporal sub-sampling operation that progressively computes a coarsened representation of the input video.
Starting with high resolution input video features, our architecture progressively aggregates the input on the temporal axis, moving from a local view of the video to a more high-level representation, as shown in Fig.~\ref{fig:architecture}.
We refer to this architecture as \ours, as it extends \ourscvpr to deal with different time granularities thanks to its hierarchical processing.

\subsubsection{Representing videos as graphs.}
A video $\mathcal{V}$ can be seen as a dense sequence of $N$ fixed-length temporal segments encoded as $\mathbf{x} = \{\mathbf{x}_1, \mathbf{x}_2, \dots, \mathbf{x}_N \}$, where $\mathbf{x}_i \in \mathbb{R}^D$ represents the features of the corresponding segment $v_i$ computed using a video features extractor $\mathcal{F}$, \eg EgoVLP~\cite{lin2022egocentric}.
The video can be interpreted as a graph~$\mathcal{G}$:
\begin{equation}
    \mathcal{G} = (\mathbf{X}, \mathcal{E}, \mathbf{pe})
\end{equation}
where $\mathbf{X} \in \mathbb{R}^{N \times D}$ is a matrix encoding the features of the graph nodes in its rows,
edge~$e_{ij} \in \mathcal{E}$ connects nodes~$i$ and~$j$ with a temporal distance considered relevant when lower than a threshold $\tau$ and the attribute $\mathbf{pe} \in \mathbb{R}^{N}$ encodes the \emph{timestamp} (in seconds).
Encoding videos as graphs enables the use of graph neural networks to learn the complex temporal relations between video segments and to cast different egocentric vision tasks as operations on these graphs. The proposed architecture is built on three components:
\begin{enumerate}
    \item a \emph{temporal} backbone $\mathcal{M}_{t}$, which uses a stack of TDGC layers and subsampling operations to implement hierarchical temporal reasoning;
    \item a set of \emph{task-specific projection necks}~$\mathcal{N}_k$ mapping the node embeddings to the features space of task $\mathcal{T}_k$;
    \item a set of \emph{task-specific heads}~$\mathcal{H}_k$ that map to the output space of each task.
\end{enumerate}
Let~$\mathcal{G}^{(0)}$ represent the initial graph of the input video~$\mathcal{V}$, where each node's position $\mathbf{pe}$ is initialized to the midpoint of the corresponding video segment.
At each stage~$l$, the \emph{temporal} backbone $\mathcal{M}_{t}$ performs temporal aggregation on the input graph $\mathcal{G}^{(l)}$ and outputs an updated graph $\mathcal{G}^{(l+1)}$. This is done using a sequence of TDGC layers and temporal subsampling operations to progressively enlarge the temporal extent of the nodes
while reducing the nodes cardinality of the graph.
Subsampling is implemented as a mean/max pooling operation over each node and its neighbors, then removing every alternate node, halving the total number of nodes.
The edges of the graph are recomputed accordingly by scaling the position of each node by a factor $2^l$, where $l$ is the index of the stage of the hierarchical temporal backbone.
Overall, the output of the temporal backbone $\mathcal{M}_{t}$ maps the input graph~$\mathcal{G}^{(0)}$ to a set of graphs:
\begin{equation}
    \mathcal{M}_{t}: \mathcal{G}^{(0)} \to \{\mathcal{G}^{(1)}, \mathcal{G}^{(2)}, \dots, \mathcal{G}^{(L)}\},
\end{equation}
where $L$ is the total number of stages in the backbone and each graph $\mathcal{G}^{(\cdot)}$ is a progressively coarsened representation of the input video. The number of stages $L$ depends on the task: for fine-grained tasks, \eg AR or OSCC, a single stage is enough, while we use multiple stages for tasks that reason over a longer horizon. More details are reported in Sec.~\ref{sec:exp_impl_details}.
The architecture of the \emph{temporal} backbone is shown in Fig.~\ref{fig:architecture}.

\subsubsection{Temporal Distance Gated Convolution (TDGC).}
\begin{figure}[t]
    \centering
    \includegraphics[trim=0.3cm 0.3cm 0.3cm 0,width=0.98\columnwidth]{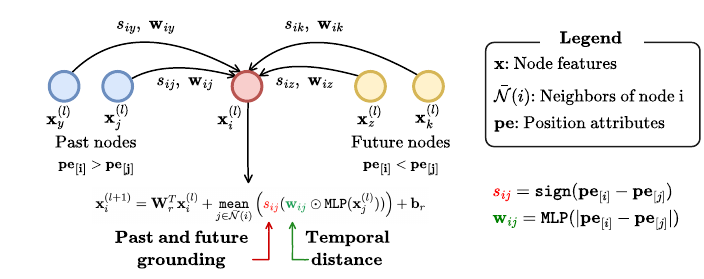}
    \caption{\textbf{Temporal Distance Gated Convolution layer (TDGC)}, specifically designed to integrate \emph{past and future events grounding} ($s_{ij}$) and to \emph{reason about the temporal distance} between nodes ($\mathbf{w}_{ij}$) in the aggregation step.}\label{fig:tdgc}
\end{figure}
Each stage of the \emph{temporal} backbone $\mathcal{M}_{t}$ is built as a stack of~$N_l$ GNN layers, which we call Temporal Distance Gated Convolution (TDGC).
These layers are designed to preserve and encode the temporal sequence of information, capturing the relative past and future dependencies between nodes. The proposed graph convolution layer, visualized in Fig.~\ref{fig:tdgc},
is explicitly designed to incorporate the relative positions between the root node and its neighbors in the message passing step.
More specifically, given two nodes $i$ and $j$ at layer $l$, we compute $s_{ij}$ as the sign of the relative temporal distance between the nodes and $\mathbf{w}_{ij}$ as a learnable projection of their relative distance (in absolute value):
\begin{equation}
    s_{ij} = \mathtt{sign}(\mathbf{pe}_{[i]}^{(l)} - \mathbf{pe}_{[j]}^{(l)}), \;\;\; \mathbf{w}_{ij} = \mathtt{MLP}(|\mathbf{pe}_{[i]}^{(l)} - \mathbf{pe}_{[j]}^{(l)}|).
\end{equation}
These two factors are used to re-weight the contribution of each node $j$ in the aggregation step, as follows:
\begin{align}
    \mathbf{x}_j^{'}     & = \mathtt{MLP}\left(\mathbf{x}_j^{(l)}\right) = \phi(\mathbf{W}_n^T \mathbf{x}_j^{(l)} + \mathbf{b}_n),                                                  \\
    \mathbf{x}_i^{(l+1)} & = \mathbf{W}^T_r\mathbf{x}_i^{(l)} + \mean_{j \in \bar{\mathcal{N}}(i)} \left( s_{ij} ( \mathbf{w}_{ij}  \odot \mathbf{x}_j^{'}) \right) + \mathbf{b}_r,
\end{align}
where $\mathbf{x}_i^{(l)}$ are the features of the node $i$ at layer $l$, $\bar{\mathcal{N}}(i)$ is the set of neighbors of node $i$, $\mathbf{W}_n$, $\mathbf{W}_r$ and $\mathbf{b}_n$, $\mathbf{b}_r$ are learnable weights and biases respectively. Subscript $r$ refers to the contribution of the root node.
Our TDGC layer is inspired by previous works on Temporal Action Localization which used 1D temporal convolution~\cite{zhao2021video,zhang2022actionformer}.
However, unlike common 1D convolutions, TDGC employs shared weights to aggregate past and future nodes,
enabling its application to video segments of arbitrary length and to graphs in which the relative temporal distance between nodes is not fixed.

\subsection{Task-specific components}\label{sec:method_ts}
The temporal backbone~$\mathcal{M}_t$ is shared between all downstream tasks and is designed to support task-agnostic temporal reasoning over a stream of fixed-length video segments.
After the backbone, we attach a separate neck~$\mathcal{N}_k$ for each task~$\mathcal{T}_k$ to project the node embeddings into the feature space of the corresponding task and possibly aligning them to the temporal boundaries of the task.
Features $\mathbf{X}^{(l)}$ from the temporal backbone are first projected with the task neck~$\mathcal{N}_k$, implemented as a two-layers MLP, to obtain~$\mathbf{X}^{(l)}_k$:
\begin{equation}\label{eq:task-specific-features}
    \mathbf{X}^{(l)}_k = \mathcal{N}_k \left( \mathbf{X}^{(l)} \right) \;\;\;\text{with}\;\;\;\mathcal{N}_k: \mathbb{R}^D \to \mathbb{R}^D.
\end{equation}
The neck is shared for all the output graphs of the \emph{temporal} backbone.
Then, for tasks defined on input segments with known temporal boundaries, \eg Action Recognition, we align the node embeddings with the task annotations.
For each video segment~$v_i \in \mathcal{V}$ annotated for the task $\mathcal{T}_k$, we aggregate the node embeddings that are between the start~$s_i$ and end~$e_i$ boundaries of the segment to obtain~$\mathbf{F}_{k,[i]}^{(l)}$:
\begin{equation}
    \mathbf{F}_{k,[i]}^{(l)} = \mathtt{align} (\mathbf{X}^{(l)}_k, s_i, e_i) = \mean_{j: \;s_i<\mathbf{p}^{(l)}_{[j]}<e_i} \mathbf{X}^{(l)}_{k,[j]},
\end{equation}
where~$i$ and~$j$ are row-indices and $\mathbf{F}_{k,[i]}^{(l)}$ are the task-specific features of segment $v_i$ of the video for task $\mathcal{T}_k$.
Other tasks, \eg Temporal Action Localization, operate on the full video and do not require task-specific alignment.
In such case, the task-specific features~$\mathbf{F}_{k}^{(l)}$ are set equal to the output of the task-specific neck $\mathbf{X}^{(l)}_k$.

\subsection{Building a backpack of reusable skills}\label{sec:method_egopack}
\begin{figure}
    \centering
    \includegraphics[trim=0 0cm 1cm 0cm,width=.95\columnwidth]{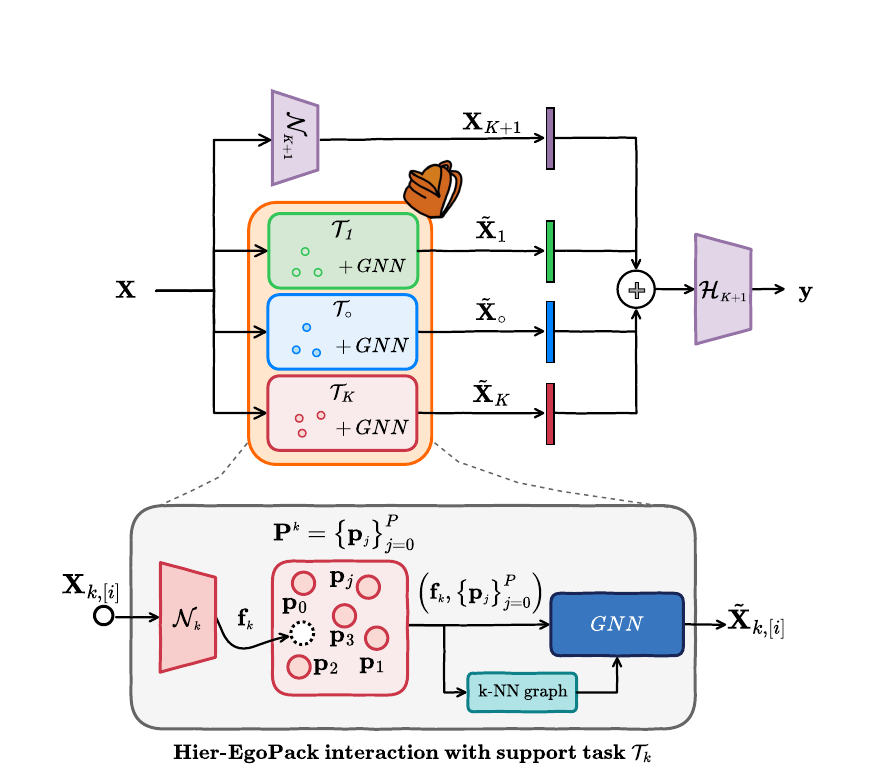}
    \vspace{-.25cm}
    \caption{
        \textbf{Learning a novel task with a backpack.}
        After the Multi-Task training phase, we extract a set of prototypes $\mathbf{P}^k$ that summarize what the network has learned from each \emph{support task} $\mathcal{T}_k$, like a backpack of skills that we can carry over.
        In this \emph{Cross-Tasks Interaction} phase, the network can peek at these different task-perspective to enrich the learning of the novel task.
    }\label{fig:interaction}
\end{figure}
To solve the \emph{novel task} $\mathcal{T}_{K+1}$, the naive approach would be to finetune the model, adding new task-specific neck $\mathcal{N}_{K+1}$ and head $\mathcal{H}_{K+1}$ and possibly updating the temporal backbone $\mathcal{M}_{t}$.
However, finetuning may not fully leverage the insights from other tasks as it could result in the loss of the previously acquired knowledge, as the model adapts to the new task.
Instead, we explicitly model the perspectives of the \emph{support tasks}, \ie the set of tasks the model has learned in the MTL pre-training step, as a set of task-specific prototypes that can be accessed by the novel task.
This approach was originally proposed as part of \ourscvpr\cite{egopack} and we provide an overview in Fig.~\ref{fig:interaction}.
We collect these task-specific prototypes from videos annotated for action recognition, as human actions can be seen as the common thread behind the different tasks.

Practically, we forward these action samples through the temporal backbone, align them based on the action recognition annotations and project their features using the task-specific necks $\mathcal{N}_k$ of each task to obtain the task-specific features $\mathbf{F}_k$ for each task in the MTL pre-training phase. Each row in $\mathbf{F}_k$ encodes the perspective of each task for the same video segment.
To summarize these features into prototypes we aggregate them according to the action label of the corresponding action segment, \ie, a \textit{verb} and \textit{noun} pair:
\begin{equation}
    \mathbf{P}^k = \{ \mathbf{p}^{k}_{0}, \mathbf{p}^{k}_{1}, \dots, \mathbf{p}^{k}_{P
    } \} \in \mathbb{R}^{P \times D},
\end{equation}
for each task $\mathcal{T}_k$, where $P$ is the number of unique \textit{(verb, noun)} pairs in the dataset and $D$ is the size of the task-specific features.
These prototypes are frozen and represent a \emph{summary} of what the models has learned during the multi-task pre-training process, creating an abstraction of the gained knowledge. They can be then reused when learning a \emph{novel task}, like a backpack of skills that the model can carry over.
Notably, storing the model's knowledge in the prototypes allows for fine-tuning the temporal backbone, which is especially valuable when the novel task has a different temporal granularity compared to the previous tasks.

\subsection{Learning a novel task with a backpack}\label{sec:method_egopack_learning}
Let us now consider the case in which we want to solve a novel task $\mathcal{T}_{K+1}$.
The model can exploit the perspective of the previously seen tasks by comparing the output of the task-specific necks for tasks $\mathcal{T}_{1,\dots,K}$ with their corresponding prototypes.
When learning the novel task $\mathcal{T}_{K+1}$, the output graphs of the \emph{temporal} backbone are forwarded through all projection necks to obtain the task-specific features $\mathbf{X}^{(l)}_{k}$, as defined in Eq.~\ref{eq:task-specific-features}.
To improve readability we hereinafter omit the superscript indicating the specific stage $l$ at the temporal backbone stage.
These features are used as \emph{queries} to match the corresponding task prototypes $\mathbf{P}^k$, using $k$-NN in the features space to look for the closest prototypes.
Task features and their neighboring prototypes form a \emph{graph-like} structure, on which message passing is performed to enrich the task-specific features $\mathbf{X}^{(l)}_k$, following an iterative refinement approach, using $M$ layers of SAGE convolution.

At each layer $m$ of \ours, we update the features~$\mathbf{X}_{k,[i]}$ from stage $l$ of the temporal backbone by combining them with its closest prototypes~$\bar{\mathcal{N}}(i)$:
\begin{equation}
    \mathbf{X}^{(m+1)}_{k,[i]} = \mathbf{W}^{(m)}_{r} \mathbf{X}^{(m)}_{k,[i]} + \mathbf{W}^{(m)} \cdot  \mean_{\mathbf{p}^k_{j} \, \in \, \bar{\mathcal{N}}(i)} \mathbf{p}^k_{j},
\end{equation}\label{eq:egopackgnn}
where~$\mathbf{p}^k_{j} \, \in \, \bar{\mathcal{N}}(i)$ are the \emph{activated prototypes} for the given task, \ie the set of closest task-specific prototypes in $\mathbf{P}^k$ with respect to $\mathbf{X}_{k,[i]}$, and $\mathbf{W}^{(m)}_{r},\mathbf{W}^{(m)}$ are learnable projections of the input features and the aggregated neighbors, respectively. Eq.~\ref{eq:egopackgnn} is applied to features from all $l$ stages of the hierarchical temporal backbone.
Notably, only the task features are refined while the task prototypes remain frozen to preserve the original perspectives seen by the network.
We denote the output of this interaction process as $\tilde{\mathbf{X}}^{(l)}_k$.
These features are then possibly aligned to the boundaries of the novel task to obtain $\tilde{\mathbf{F}}^{(l)}_k$, as discussed in Sec.~\ref{sec:method_ts}.

In this process, the \emph{task-specific necks} of the support tasks $\mathcal{N}_{1,\dots,K}$ are initialized from the multi-task training and updated during the task-specific finetuning process, allowing the model to explore the set of task prototypes and to select the most informative ones for each input sample.
Moreover, to allow the model to learn complementary cues specific to the novel task, we add a new pair of neck $\mathcal{N}_{K+1}$ and head $\mathcal{H}_{K+1}$.
We evaluate different fusion strategies to integrate the novel task with the perspectives gained from the previous tasks.
In \emph{features-level} fusion, we average the task-specific features for the novel task $\mathbf{F}_{K+1}$ with the \textit{refined} perspectives from the previous tasks~$\tilde{\mathbf{F}}_k$.
In \emph{logits-level} fusion, we keep a set of separate heads, one for each task $\mathcal{T}_{1,\dots,K}$, feed the features $\tilde{\mathbf{F}}_k$ to each head separately and sum their outputs, as in the original \ourscvpr implementation.
Intuitively, this approach allows each task to cast a vote on the final prediction, based on its perspective on the same video segment.

\subsection{Training process}
We train our models using only supervision of the known task, for both single and multi-task models. More details are reported in Sec.~\ref{sec:exp_impl_details}.
When training \ours, we finetune the \emph{temporal} backbone, the task-specific projection necks and the heads. Gradient updates from the support tasks are not propagated to the \emph{temporal} backbone.
\section{Experiments}\label{sec:experiments}
We first introduce in Sec.~\ref{sec:exp_setting} the tasks addressed in this work and the implementation details for our models and the \emph{Task-Translation} baseline in Sec.~\ref{sec:exp_impl_details}.
We report quantitative results for \ours in Sec.~\ref{sec:exp_quantitative}, evaluate different design choices in Sec.~\ref{sec:exp_ablations} and demonstrate the effectiveness of our approach on the test-set in Sec.~\ref{sec:exp_benchmarks}.
Finally, in Sec.~\ref{sec:exp_qualitative} we show qualitative results demonstrating the interaction process of \ours.

\subsection{Setting}\label{sec:exp_setting}
We validate our approach on \egofourd~\cite{ego4d}, a large scale dataset with 3.6k hours of egocentric videos capturing unscripted daily-life human activities, focusing on five \egofourd benchmarks that cover different temporal granularities.
\emph{Fine-grained tasks} focus on short-term understanding of the video, usually a few seconds long, and include:
\begin{itemize}
    \item \emph{Action Recognition (AR)}: given a video segment, predict the verb and noun action labels describing the interaction from a taxonomy of 115 and 478 verb and noun classes respectively. We report verb and noun top-1 accuracy.\footnote{This task is not an official \egofourd~\cite{ego4d} task and was initially introduced by EgoT2~\cite{egot2} using the LTA annotations.}
    \item \emph{Object State Change Classification (OSCC)}: given a video segment, predict the presence (or absence) of an object state change, \eg a glass being filled (transition from \textit{empty} to \textit{full}). We report accuracy.
    \item \emph{Point of No Return (PNR)}: given a video segment containing an object state change, predict the temporal frame when the change happens. Predictions are evaluated using the absolute temporal distance from the ground truth.
    \item \emph{Long Term Anticipation (LTA)}: given a video segment, predict the sequence of Z future actions (verb and noun label pairs) the camera wearer is likely to perform next. Performance is measured in terms of verbs and nouns Edit Distance (ED) between the predicted sequence and the ground truth, for the best sequence out of K predictions. In \egofourd, $Z=20$ and $K=5$.
\end{itemize}
Other tasks may require both short and long term understanding of the input video.
Among these, we analyze the \emph{Moment Queries (MQ)} task, which requires predicting the set of activities performed in the video among 110 labels with the corresponding start and end timestamps.
For all tasks, we use the version \textit{v1} of the annotations.

\subsection{Implementation Details}\label{sec:exp_impl_details}
\ours is built using pre-extracted features from fixed-size video segments. In all experiments the backbone used for feature extraction is kept frozen.
We use EgoVLP features pretrained on EgoClip~\cite{lin2022egocentric} and extracted using a window of 16 consecutive frames with an equivalent stride. EgoVLP features have size 256.
For comparison with \ourscvpr in Table~\ref{tab:cvpr}, we use Omnivore Video Swin-L~\cite{omnivore} features pre-trained on Kinetics-400~\cite{quo_vadis}, released as part of \egofourd~\cite{ego4d} and extracted using dense sampling over a window of 32 frames with a stride of 16 frames and features size 1536.
In principle, \ours is agnostic to the features extractor and could adopt other architectures.
We train all the single, multi-task and \ours models for 15 epochs, using the Adam optimizer. Learning rate is set to $1\mathrm{e}{-4}$ for all tasks, with the exception of the OSCC and PNR tasks which use $1\mathrm{e}{-5}$, and follow a cosine annealing schedule with a linear warmup of 5 epochs.
We repeat our experiments three times with different random seed and report the average performance.
All tasks share the same temporal and cross-task interaction architecture, with minimal task-specific hyper-parameter tuning.
The task prototypes are built using samples from the train split of the AR~task.

\begin{table*}[tb]
  \centering
  \footnotesize
  \caption{\ours on \egofourd Human-Object Interaction (HOI) and Moment Queries (MQ) tasks.}\label{tab:main_results}
  \vspace{-0.25cm}
  \begin{tabularx}{1.0\textwidth}{Xcccccccc}
    \toprule

                                 & \multicolumn{2}{c}{\textbf{AR}} & \multicolumn{1}{c}{\textbf{OSCC}} & \multicolumn{2}{c}{\textbf{LTA}} & \multicolumn{1}{c}{\textbf{PNR}} & \multicolumn{1}{c}{\textbf{MQ}}                                                         \\
    \cmidrule(lr){2-8}

                                 & \textbf{Verbs Top-1 (\%)}       & \textbf{Nouns Top-1 (\%)}         & \textbf{Acc. (\%)}               & \textbf{Verbs ED ($\downarrow$)} & \textbf{Nouns ED ($\downarrow$)} & \textbf{Loc. Err. ($\downarrow$)} & \textbf{mAP}     \\

    \midrule

    Ego4D Baselines~\cite{ego4d} & 22.18                           & 21.55                             & 68.22                            & 0.746                            & 0.789                            & \underline{0.62}                  & 6.03             \\
    EgoT2s~\cite{egot2}          & 23.04                           & 23.28                             & 72.69                            & 0.731                            & 0.769                            & \textbf{0.61}                     & N/A              \\

    %\midrule
    \ourscvpr~\cite{egopack}     & 25.10                           & 31.10                             & 71.83                            & 0.728                            & 0.752                            & \textbf{0.61}                     & N/A              \\

    \midrule
    \midrule

    Single Task                  & \underline{26.93}               & 33.50                             & 75.22                            & \underline{0.728}                & 0.752                            & \underline{0.62}                  & 20.2             \\
    MTL                          & 26.31                           & \underline{33.90}                 & 74.79                            & 0.730                            & 0.754                            & \underline{0.62}                  & 18.5             \\
    MTL + FT                     & 26.71                           & 33.51                             & 75.00                            & 0.728                            & 0.749                            & \textbf{0.61}                     & 19.9             \\
    MTL + HT                     & 26.07                           & 33.20                             & 74.27                            & 0.729                            & 0.748                            & \underline{0.62}                  & N/A              \\
    \midrule
    Task-Translation$^\dagger$   & 26.10                           & 33.83                             & \textbf{76.42}                   & 0.729                            & \underline{0.750}                & 0.63                              & \underline{20.5} \\
    \textbf{\ours}               & \textbf{27.30}                  & \textbf{34.65}                    & \underline{75.60}                & \textbf{0.725}                   & \textbf{0.741}                   & \textbf{0.61}                     & \textbf{21.0}    \\

    \bottomrule
  \end{tabularx}
  \begin{tablenotes}
    \scriptsize
    \item \emph{Single Task} uses the same hierarchical GNN-based architecture to model all tasks, with minimal task-specific differences. \emph{Multi-Task Learning (MTL)} uses hard parameter sharing to jointly learn all tasks, which may result in negative transfers. \emph{Ego-T2s}~\cite{egot2} learns to translate features across tasks to optimize the primary task. \emph{\ours} builds on the unified architecture of the Temporal Graph and learns to exploit the perspective of different tasks for efficient knowledge transfer to the novel task. Performances of \ours are evaluated over three runs with different random seeds using accuracy for AR and OSCC, Edit Distance for LTA, temporal localization error (in seconds) for PNR and mAP for MQ. $^\dagger$\emph{Task-Translation} implements the same cross-task translation mechanism of EgoT2s~\cite{egot2} using a frozen EgoVLP backbone, as for \ours. Best results are reported in bold, second best are underlined.
  \end{tablenotes}
  \vspace{-3mm}
\end{table*}

\subsubsection{Task-specific design choices.}\label{ts-design}
\ourscvpr constructs the input graph differently based on the task, \ie each action or sub-segment is mapped to a different node in AR or OSCC respectively, which may result in inconsistencies in how segments with different temporal granularities are processed by the temporal backbone.
On the contrary, with \ours we standardize the graph construction process for all tasks.
Specifically, features from fixed-length segments are extracted densely from the entire video and each segment is mapped to a node of the graph.
Temporal reasoning is performed on these \textit{dense} temporal graphs, followed by a task-specific projection~$\mathcal{N}_k$ and an optional alignment step.
Depending on the temporal granularity of the downstream task, we take the output processed graphs of the temporal model after the first stage $\mathcal{G}^{(1)}$ (\textit{fine-grained tasks}) or from all the stages $\{\mathcal{G}^{(1)}, \mathcal{G}^{(2)}, \dots, \mathcal{G}^{(L)}\}$ (\textit{variable-resolution tasks}).
For tasks in which temporal boundaries are known, features within the boundaries are averaged to obtain a single instance-level embedding as input for the task-specific neck. With the exception of MQ, all tasks fall into this category.
For MQ, we predict an action for each segment in the input video and use Non-Maximum Suppression (NMS) to filter predictions, consistently with previous approaches. For NMS, we use the same configuration as ActionFormer~\cite{zhang2022actionformer} and set the $\sigma$ parameter to $2.0$, which was empirically found to reduce the penalty on \textit{near-replicate} predictions~\cite{sui2023nmsthresholdmattersego4d}. Therefore, no specific alignment is needed for this task.

Task-specific necks are implemented as two-layers MLPs. The heads are also implemented as multi-layer projections that map to the output space of the task, with the exception of the LTA task. In this case, we first build \textit{on-the-fly} a graph with $K$ nodes initialized to the output of the temporal model, where K is the number of future actions to predict. We then process this graph with a two layers TDGC, before feeding the features to the verb and noun classifiers.

AR, OSCC and LTA are trained with standard cross entropy loss, while PNR uses binary cross entropy. The classification and regression heads of the MQ task are trained with the focal~\cite{ross2017focal} and DIoU~\cite{zheng2020distance} losses respectively, following the same protocol as ActionFormer~\cite{zhang2022actionformer} to match predictions at different scales with their temporally closest ground truth.

\subsubsection{Task-Translation baseline implementation.}
Due to the differences in the network architecture and training data employed, a comparison between \ours and EgoT2~\cite{egot2} is not straightforward.
Indeed, EgoT2's Single Task are based on SlowFast~\cite{slowfast} for AR and LTA, I3D ResNet-50~\cite{carreira2017quo} for OSCC and PNR and VSGN~\cite{zhao2021video} for MQ.
These models are end-to-end trained on the benchmarks' data, unlike \ours which relies on pre-extracted features and does not train the video feature extractor.
Therefore, we introduce a comparable baseline, which we call \emph{Task Translation}, by adapting the cross-task translation mechanism of EgoT2 to our setting.
As in EgoT2s, \emph{Task Translation} learns a transformer encoder on top of the Single Task models to combine the perspective of the different tasks.
Furthermore, EgoT2 supports only tasks with homogeneous temporal granularity.
With \emph{Task Translation}, we extend the translation mechanism to support tasks with different temporal granularities and include in this analysis the same tasks as \ours.

Formally, \emph{Task Translation} combines a set of $K$ Single Task models trained independently.
Each Single Task model outputs a sequence of $N_k$ task-specific tokens $\mathbf{F}_{k} = [\mathbf{f}^{1}_k, \mathbf{f}^{2}_k, \dots, \mathbf{f}^{N_k}_{k}]$ with $\mathbf{f}^{i}_{k} \in \mathbb{R}^{D}$, along with the position attribute $\mathbf{pe}_{k}\in\mathbb{R}^{N_k}$, as defined in Sec.~\ref{sec:method}.
Task-specific tokens and the position attribute are concatenated on the sequence dimension to obtain the full features $\mathbf{F} \in \mathbb{R}^{N \times D}$ and position attribute $\mathbf{pe} \in \mathbb{R}^{N}$, where $N$ is the total number of tokens across all the tasks. We define the \emph{Task Translation} operation as $\tilde{\mathbf{F}} = \mathtt{ENC}(\mathbf{F}, \mathbf{A})$, where $\mathbf{A}$ is a binary attention mask defined as:
\begin{equation}
    \mathbf{A}_{[ij]} = \begin{cases}
        1 & \quad\text{if}\;\left| \mathbf{pe}_{[i]} - \mathbf{pe}_{[j]} \right| \le 2^{l} \\
        0 & \quad\text{otherwise}                                                          \\
    \end{cases},
\end{equation}
where $l$ is the index of the stage in the hierarchical backbone that produced features $\mathbf{f}_i$. The mask restricts the self-attention operation to tokens that are within the same temporal window.
We parameterize $\mathtt{ENC}$ as a transformer encoder with $l$ layers and $h$ attention heads, with the same output size as the input features.
Finally, we take the slice of the transformer output $\tilde{\mathbf{F}}$ corresponding to the features of the primary task and forward them through the task-specific~head.

\subsection{Quantitative results}\label{sec:exp_quantitative}
We show the main results of \ours in Table~\ref{tab:main_results}, comparing our approach with the \egofourd baselines~\cite{ego4d}, the task-translation framework EgoT2~\cite{egot2} and the previous iteration of our work \ourscvpr~\cite{egopack}.

We proceed incrementally from the \emph{Single Task} models, \ie each task is trained separately using our GNN-based hierarchical architecture.
Conversely, \egofourd baselines and EgoT2 use SlowFast~\cite{slowfast} for AR and LTA, I3D ResNet-50~\cite{carreira2017quo} for OSCC and PNR and VSGN~\cite{zhao2021video} for MQ, with different configuration and hyper-parameters for each task.
In contrast, our \emph{Single Task} models employ the same architecture and a pair of task-specific neck and head.
\emph{Multi-Task Learning} (MTL) baselines are built following the same approach, \ie sharing the same architecture across all the tasks.
In this setting, we observe suboptimal performance in some tasks, particularly in AR (Verb), OSCC, and MQ. We attribute this to potential negative transfer effects.
We also consider a \emph{MTL+FT} baseline in which the MTL model is finetuned on the novel task, and  \emph{MTL+HT} which takes the frozen temporal backbone from the MTL training and learns new task-specific neck~$\mathcal{N}_K$ and head~$\mathcal{H}_K$ for the novel task.
These baselines exhibit comparable performance to the \emph{Single Task} models, showing that fine-tuning multi-task models is not the ideal approach to transfer knowledge across tasks as it does not explicitly exploit the semantic similarities and perspectives offered by different tasks.

\subsubsection{Task-Translation baseline results.}
\emph{Task-Translation} shows consistent improvements compared to both Single Task and Multi-Task models, with the sole exception of AR.
These results prove the effectiveness of the cross-task translation mechanism and show that different tasks learn representations that are partially complimentary to each other.
However, we remark the \emph{Task-Translation} mechanism is inefficient by design as it requires different models for each supported task.
Each single task model in the ensemble looks at a different perspective for the same input, without explicitly recalling the entire knowledge gained by the models.
In contrast, the task prototypes in \ours provide a comprehensive and easy-to-access abstraction of the model’s learned knowledge, enabling the extraction of relevant insights tailored to the specific sample and task.

\subsubsection{Comparison with \ourscvpr}\label{sec:comparisonCVPR}

\begin{table}[tb]
    \centering
    \footnotesize
    \setlength{\tabcolsep}{5pt}
    \caption{Comparison of \ourscvpr and \ours using Omnivore features.}\label{tab:cvpr}
    \vspace{-0.25cm}
    \begin{tabularx}{1.0\columnwidth}{Xcccccc}
        \toprule

                                   & \multicolumn{2}{c}{\textbf{AR}} & \multicolumn{1}{c}{\textbf{OSCC}} & \multicolumn{2}{c}{\textbf{LTA}} & \multicolumn{1}{c}{\textbf{PNR}}                                        \\

                                   & \textbf{Verbs}                  & \textbf{Nouns}                    & \textbf{Acc.}                    & \textbf{Verbs}                   & \textbf{Nouns}    & \textbf{Err.}    \\

        \midrule

        Single Task~\cite{egopack} & 24.25                           & 30.43                             & 71.26                            & 0.754                            & 0.752             & \textbf{0.61}    \\
        MTL~\cite{egopack}         & 22.05                           & 29.44                             & 71.10                            & 0.740                            & 0.746             & \underline{0.62} \\
        \ourscvpr~\cite{egopack}   & \underline{25.10}               & 31.10                             & \textbf{71.83}                   & \textbf{0.728}                   & 0.752             & \textbf{0.61}    \\

        \midrule

        Single Task                & 24.41                           & 31.41                             & 71.74                            & 0.733                            & \underline{0.743} & \textbf{0.61}    \\
        MTL                        & 23.72                           & \underline{31.43}                 & 71.33                            & 0.737                            & 0.756             & \underline{0.62} \\
        \textbf{\ours}             & \textbf{25.33}                  & \textbf{31.64}                    & \underline{71.77}                & \underline{0.729}                & \textbf{0.741}    & \textbf{0.61}    \\

        \bottomrule
    \end{tabularx}
    \begin{tablenotes}
        \scriptsize
        \item Comparison between \emph{\ourscvpr} and \emph{\ours} using the same input features (Omnivore) and tasks, \ie, AR, OSCC, LTA and PNR.
    \end{tablenotes}
\end{table}
We compare \ours with our previous iteration \ourscvpr~\cite{egopack} in Table~\ref{tab:cvpr}, using the same \emph{fine-grained} tasks, \ie AR, OSCC, LTA and PNR, and same pre-extracted features (Omnivore).
\ours, thanks to its novel GNN layer with strong temporal reasoning, has on average better performance compared to the Single Task models from the original \ourscvpr.

\subsection{Ablations}
\label{sec:exp_ablations}
\begin{table*}[tb]
    \caption{Ablations on different components of the hierarchical temporal model.}
    \label{tab:temp_gnn_ablations}
    \vspace{-0.25cm}
    \hfill
    \begin{minipage}{0.3\textwidth}
        \footnotesize
        \begin{tabularx}{1.0\textwidth}{@{}Xccc@{}}
            \toprule
            $\textbf{N}_l$ & \textbf{mAP}   & \textbf{R@1}   & \textbf{R@5}   \\
            \midrule
            1              & 18.57          & 31.47          & 53.36          \\
            \textbf{2}     & \textbf{20.21} & 34.15          & 56.78          \\
            3              & 19.75          & 35.08          & 57.23          \\
            4              & 19.93          & \textbf{35.24} & \textbf{59.28} \\
            \bottomrule
        \end{tabularx}
        \begin{tablenotes}[flushleft]
            \scriptsize
            \item \noindent Number of TDGC layers in each stage of the temporal backbone.
        \end{tablenotes}
    \end{minipage}%
    \hfill
    \begin{minipage}{0.3\textwidth}
        \footnotesize
        \begin{tabularx}{1.0\textwidth}{@{}Xccc@{}}
            \toprule
            \textbf{Pooling} & \textbf{mAP}   & \textbf{R@1}   & \textbf{R@5}   \\
            \midrule
            batch ss.        & 19.95          & 34.30          & 57.69          \\
            video ss.        & 19.35          & 33.43          & 56.53          \\
            \textbf{max}     & 19.87          & \textbf{34.41} & \textbf{58.14} \\
            mean             & \textbf{20.21} & 34.15          & 56.78          \\
            \bottomrule
        \end{tabularx}
        \begin{tablenotes}[flushleft]
            \scriptsize
            \item \noindent Pooling strategy to progressively reduce the number of nodes in the temporal backbone.
        \end{tablenotes}
    \end{minipage}%
    \hfill
    \begin{minipage}{0.3\textwidth}
        \footnotesize
        \begin{tabularx}{1.0\textwidth}{@{}Xccc@{}}
            \toprule
            \textbf{Temp. Thresh. $\tau$} & \textbf{mAP}   & \textbf{R@1}   & \textbf{R@5}   \\
            \midrule
            1                             & 18.21          & 31.09          & 54.92          \\
            \textbf{2}                    & \textbf{20.21} & \textbf{34.15} & \textbf{56.78} \\
            4                             & 19.63          & 32.49          & 54.53          \\
            8                             & 20.07          & 31.67          & 52.46          \\
            \bottomrule
        \end{tabularx}
        \begin{tablenotes}[flushleft]
            \scriptsize
            \item \noindent Temporal distance threshold to define a connection between nodes in the temporal graph.
        \end{tablenotes}
    \end{minipage}%
    \hfill
\end{table*}

\begin{table}[tb]
    \centering
    \caption{Ablations on different GNNs for the hierarchical backbone.}
    \label{tab:temp_gnn_mAP}
    \vspace{-0.25cm}
    \footnotesize
    \begin{tabularx}{1.0\columnwidth}{X|ccccc|c@{}}
        \toprule
        \multicolumn{1}{c}{}                  & \multicolumn{6}{c}{\textbf{mAP @ IoU}}                                                                                                                    \\
        \multicolumn{1}{l}{\textbf{GNN}}      & \textbf{0.1}                           & \textbf{0.2}      & \textbf{0.3}      & \textbf{0.4}      & \multicolumn{1}{c}{\textbf{0.5}} & \textbf{Avg}      \\
        \midrule
        \rowcolor{lightgray!50} \multicolumn{7}{c}{Permutation-Invariant layers}                                                                                                                          \\
        \midrule
        GCN~\cite{kipf2017semisupervised}     & 21.43                                  & 18.46             & 15.51             & 12.16             & 9.21                             & 15.35             \\
        GAT~\cite{velivckovic2018graph}       & 21.58                                  & 18.57             & 15.58             & 12.12             & 9.12                             & 15.39             \\
        SAGE~\cite{graphsage}                 & 21.95                                  & 19.00             & 15.99             & 12.49             & 9.21                             & 15.73             \\
        \midrule
        \rowcolor{lightgray!50} \multicolumn{7}{c}{Temporal-aware layers}                                                                                                                                 \\
        \midrule
        SAGE + PE.$^\dagger$~\cite{graphsage} & 25.22                                  & 21.38             & 17.82             & 13.61             & 10.28                            & 17.66             \\
        SGCN~\cite{8594922}                   & \underline{25.35}                      & \underline{22.32} & \underline{19.58} & \underline{17.03} & \underline{14.39}                & \underline{19.73} \\
        \midrule
        TDGC (w/o $s_{ij}$)                   & 21.27                                  & 18.10             & 15.35             & 12.09             & 8.79                             & 15.12             \\
        TDGC (w/o $\mathbf{w}_{ij}$)          & 24.98                                  & 21.99             & 19.55             & 16.85             & 14.25                            & 19.52             \\
        \textbf{TDGC}                         & \textbf{25.83}                         & \textbf{22.93}    & \textbf{20.17}    & \textbf{17.38}    & \textbf{14.73}                   & \textbf{20.21}    \\
        \bottomrule
    \end{tabularx}
    \begin{tablenotes}[flushleft]
        \scriptsize
        \item \noindent Comparison between TDGC and other GNN layers for the stages of the temporal backbone on the MQ task.
        $^\dagger$ A sinusoidal positional encoding is added to the nodes of the input graph.
    \end{tablenotes}
\end{table}
We evaluate different design choices for the hierarchical temporal backbone in Tables~\ref{tab:temp_gnn_ablations} and~\ref{tab:temp_gnn_mAP}, focusing on the Moment Queries (MQ) task which requires temporal reasoning at multiple granularities, thus exploiting the hierarchical architecture in its entirety.

\smallskip
\smallskip
\noindent\textbf{Number of GNN layers.}
The number of convolutional layers at each stage has a mild impact on performance, as it tends to saturate after two layers (Table~\ref{tab:temp_gnn_ablations}-left).
Increasing the number of layers expands the receptive field at each stage, a goal already achieved by our pooling and hierarchical aggregation steps.
Consequently, adding more layers appears redundant given the model’s hierarchical structure.

\smallskip
\smallskip
\noindent\textbf{Pooling strategy.} We evaluate different approaches to \textit{reduce the temporal resolution of graph nodes} between subsequent layers of the temporal model (Table~\ref{tab:temp_gnn_ablations}-middle).
The \textit{batch} strategy selects alternate nodes from the batch, without considering video boundaries, which results in some noise in the node selection process.
Differently, \textit{video} selects alternate nodes from each video separately.
The \textit{mean} and \textit{max} strategies pool features from all the neighbors of each node, corresponding to past and future segments.
On the MQ task, we observe a noticeable gap between the first two strategies that drop half the nodes and the \textit{mean} and \textit{max} strategies which operate on the neighbors of each node and can better forward task-relevant information to the next layers.

\smallskip
\smallskip
\noindent\textbf{Temporal threshold.} The $\tau$ parameter controls the number of neighbors at each node in the temporal graph, as we consider the existence of an edge  $e_{ij}$ between two nodes $i$ and $j$ only if their relative temporal distance is less than the threshold $\tau$.
We observe that small values of $\tau$ are sufficient and performance deteriorates quickly with larger values, especially in terms of recall (Table~\ref{tab:temp_gnn_ablations}-right).
Also, the use of a smaller neighborhood is compensated by the hierarchical nature of our temporal backbone.

\smallskip
\smallskip
\noindent\textbf{GNN layer.}
In Table~\ref{tab:temp_gnn_mAP}, we analyze the impact of different GNN layers on MQ performance.
At each layer, the neighborhood of a node is the set of nodes within a fixed relative distance $\tau$ from the root node, which makes the GNN operate on local temporal segments of the video.
We evaluate two different approaches: i) permutation invariant~(PI) layers, which ignore the local temporal ordering of the nodes in the neighborhood, and ii) layers that explicitly incorporate temporal grounding, \ie node ordering, into their processing.
Both strategies achieve reasonable performance.
However, the absence of temporal ordering in the approaches from the first group prevents them from properly aggregating past and future nodes, resulting in subpar performance compared to strategies that include temporal grounding.

We evaluate different strategies to add temporal grounding to the GNN layers.
The simplest approach, \emph{SAGE + PE}, adds an absolute positional encoding to the node embeddings of the input graph.
This method, already used by \ourscvpr, works well in tasks that do not require strong temporal reasoning. Despite its simplicity, it outperforms all PI approaches, underscoring the significance of precise node ordering for more \textit{temporal-aware} tasks, such as the MQ.
A more advanced strategy is SGCN~\cite{8594922}, which extends GCN by using different projections for the node embeddings corresponding to past and future segments in the neighborhood.
To design an effective GNN layer for diverse video understanding tasks, we focus on two key temporal reasoning principles: (i) the ability to distinguish between past and future nodes in the aggregation phase and (ii) the relevance of each node should depend on its relative temporal distance.
SGCN addresses the first point but does not consider the relative temporal distance of the nodes, giving the same importance to close and distant nodes. Also, past and future node embeddings are projected differently despite possibly encoding the same event.
Our intuition is that the relative temporal distance should not affect the semantic content of the nodes, and therefore their projection, but only how nodes are combined in the aggregation phase.
By using our TDGC layer we adopt the same projection for all nodes and encode the temporal distance between the nodes in the aggregation step.

To analyze the impact of the aforementioned key temporal reasoning principles, Table~\ref{tab:temp_gnn_mAP} also presents an ablation study on the design choices for our TDGC. The results clearly show a significant performance drop when the $s_{ij}$ coefficients are removed, as this prevents distinguishing between past and future nodes during aggregation.
Similarly, omitting the relative position attributes $\mathbf{w}_{ij}$, which differentiates between temporally close and distant nodes, results in suboptimal performance in the downstream MQ task.

\begin{figure*}[ht]
    \hfill
    \begin{minipage}{0.325\textwidth}
        \includegraphics[trim=0.5cm 0.5cm 0.5cm 0.5cm,height=4cm]{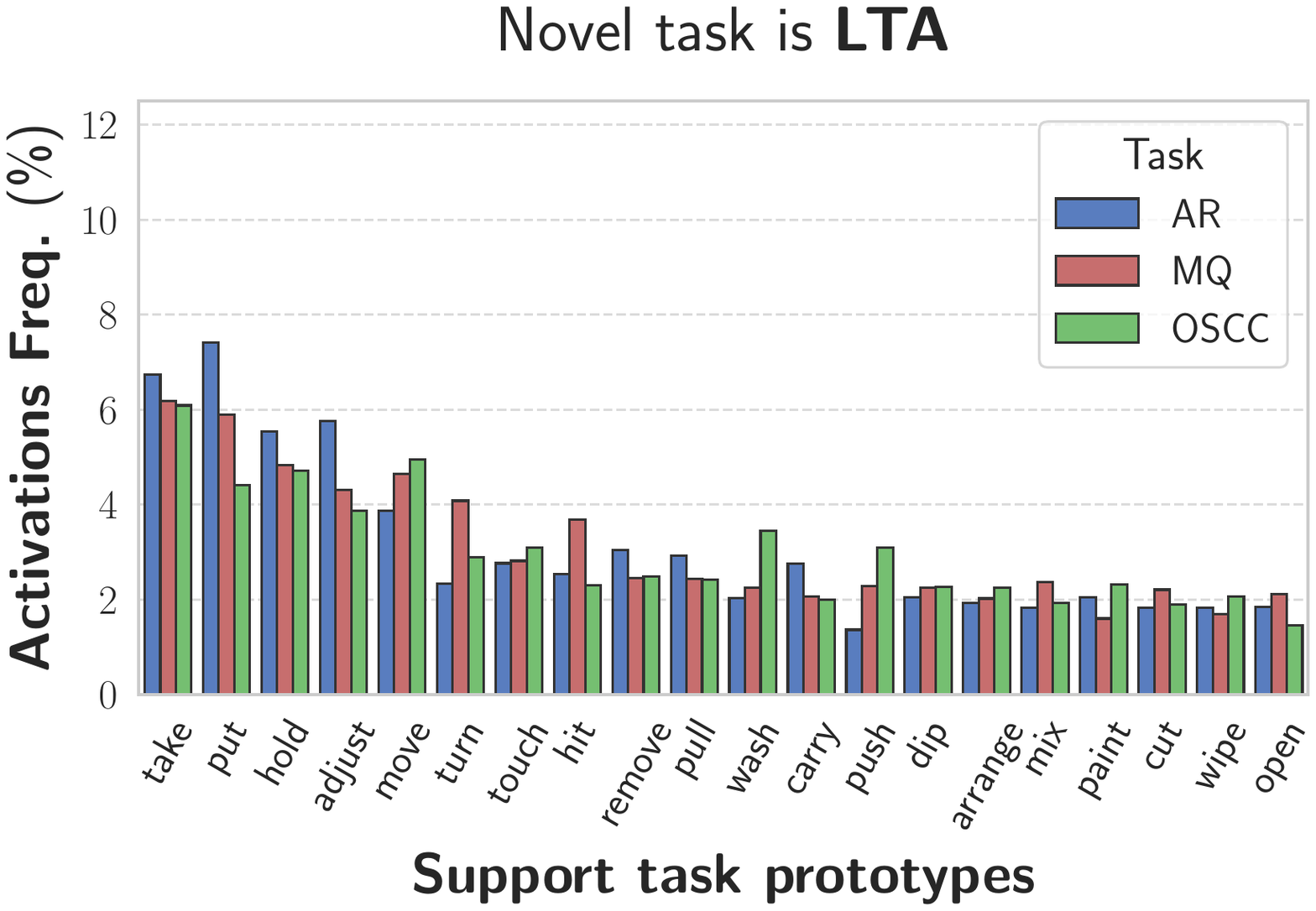}
    \end{minipage}
    \hfill
    \begin{minipage}{0.325\textwidth}
        \includegraphics[trim=0.5cm 0.5cm 0.5cm 0.5cm,height=4cm]{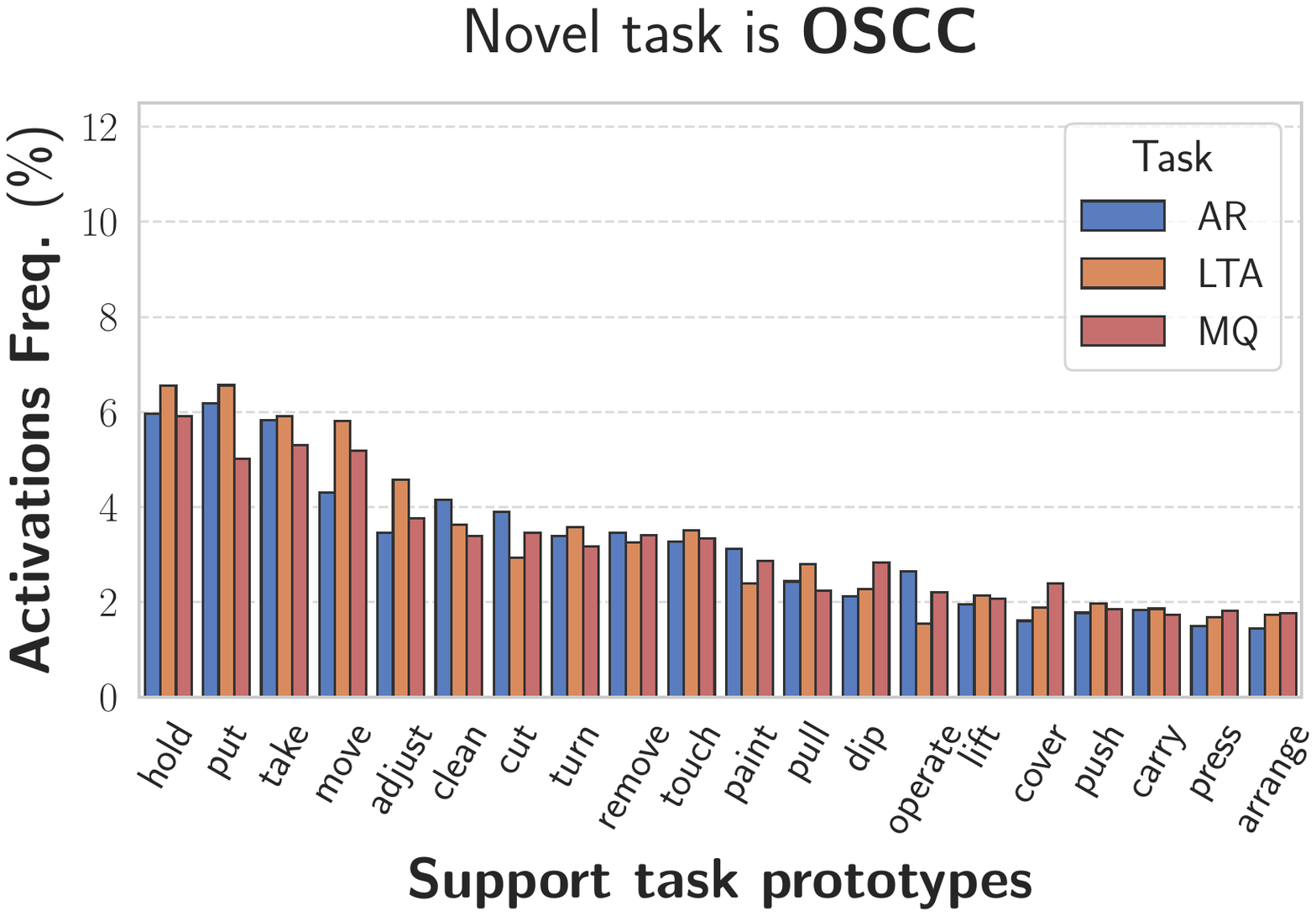}
    \end{minipage}
    \hfill
    \begin{minipage}{0.325\textwidth}
        \includegraphics[trim=0.5cm 0.5cm 0.5cm 0.5cm,height=4cm]{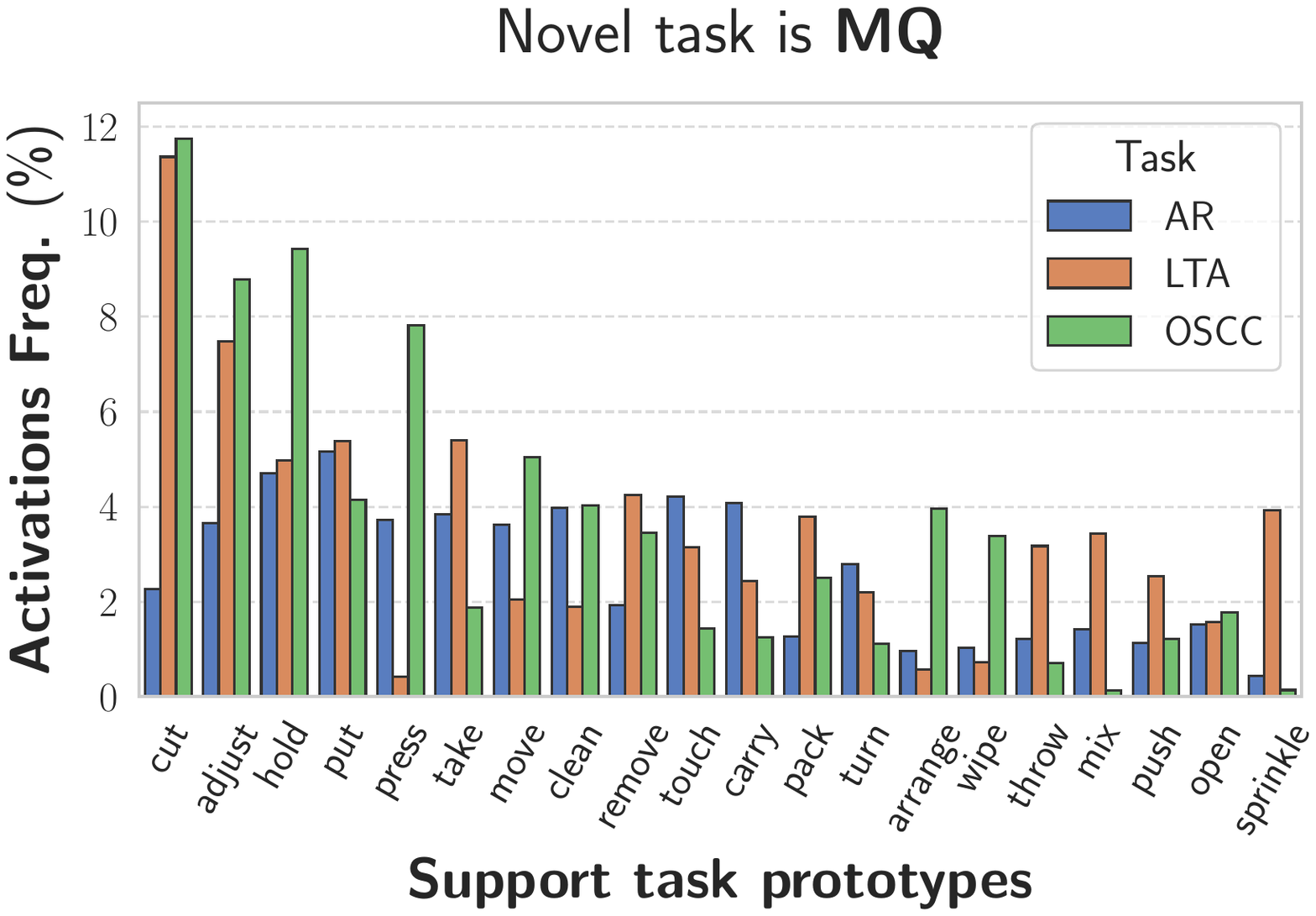}
    \end{minipage}
    \caption{\textbf{Activation frequency for the task-specific prototypes from different \emph{support tasks}}. We focus on the Top-20 most activated prototypes across the \emph{support tasks}. LTA and OSCC have more uniform activations across different support tasks, \ie, they look at similar prototypes, while MQ exhibit more diverse activations.}\label{fig:activations}
\end{figure*}
\begin{figure*}[t]
    \hfill
    \begin{minipage}{0.2425\textwidth}
        \includegraphics[trim=0.5cm 0.5cm 0.5cm 0.5cm,height=4cm]{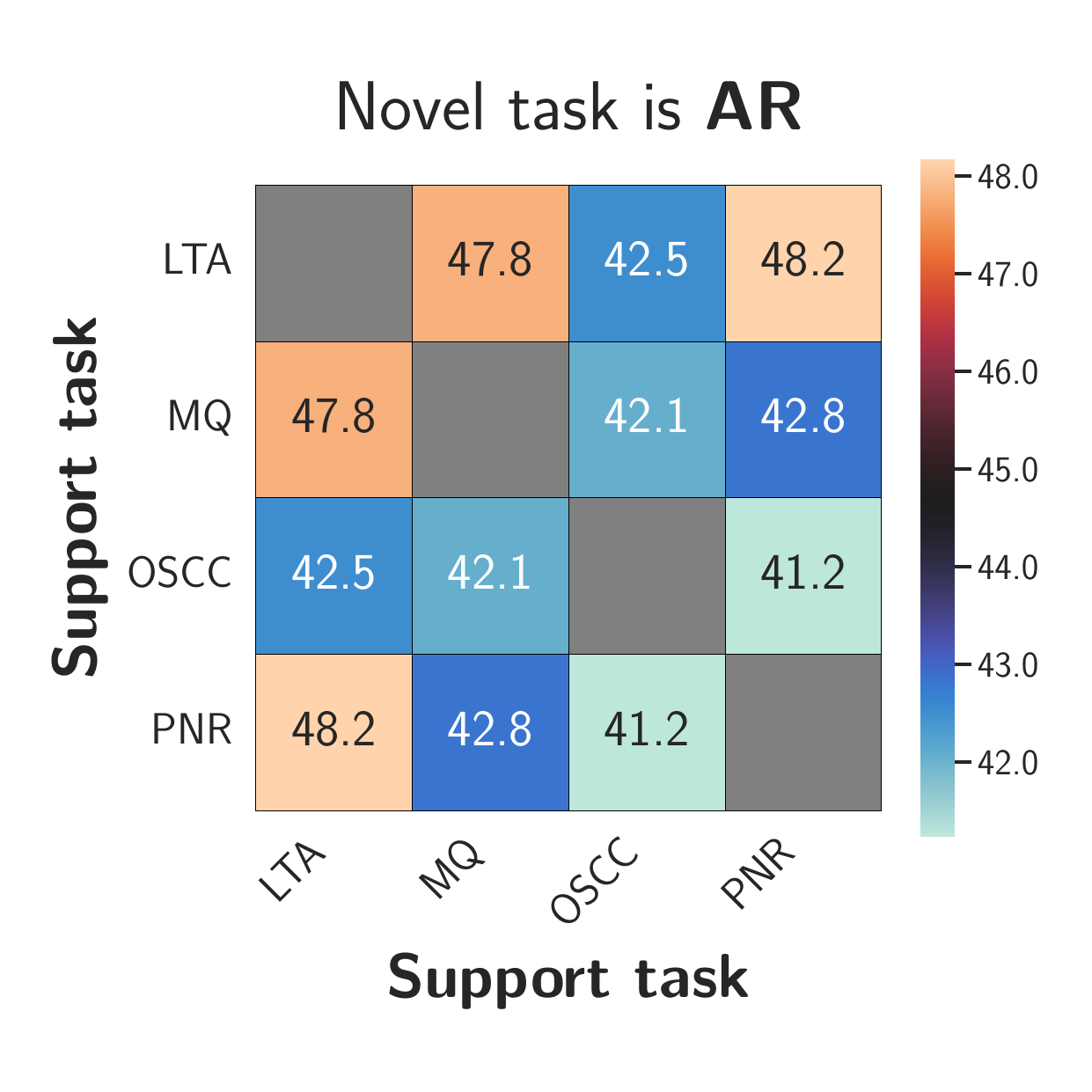}
    \end{minipage}
    \hfill
    \begin{minipage}{0.2425\textwidth}
        \includegraphics[trim=0.5cm 0.5cm 0.5cm 0.5cm,height=4cm]{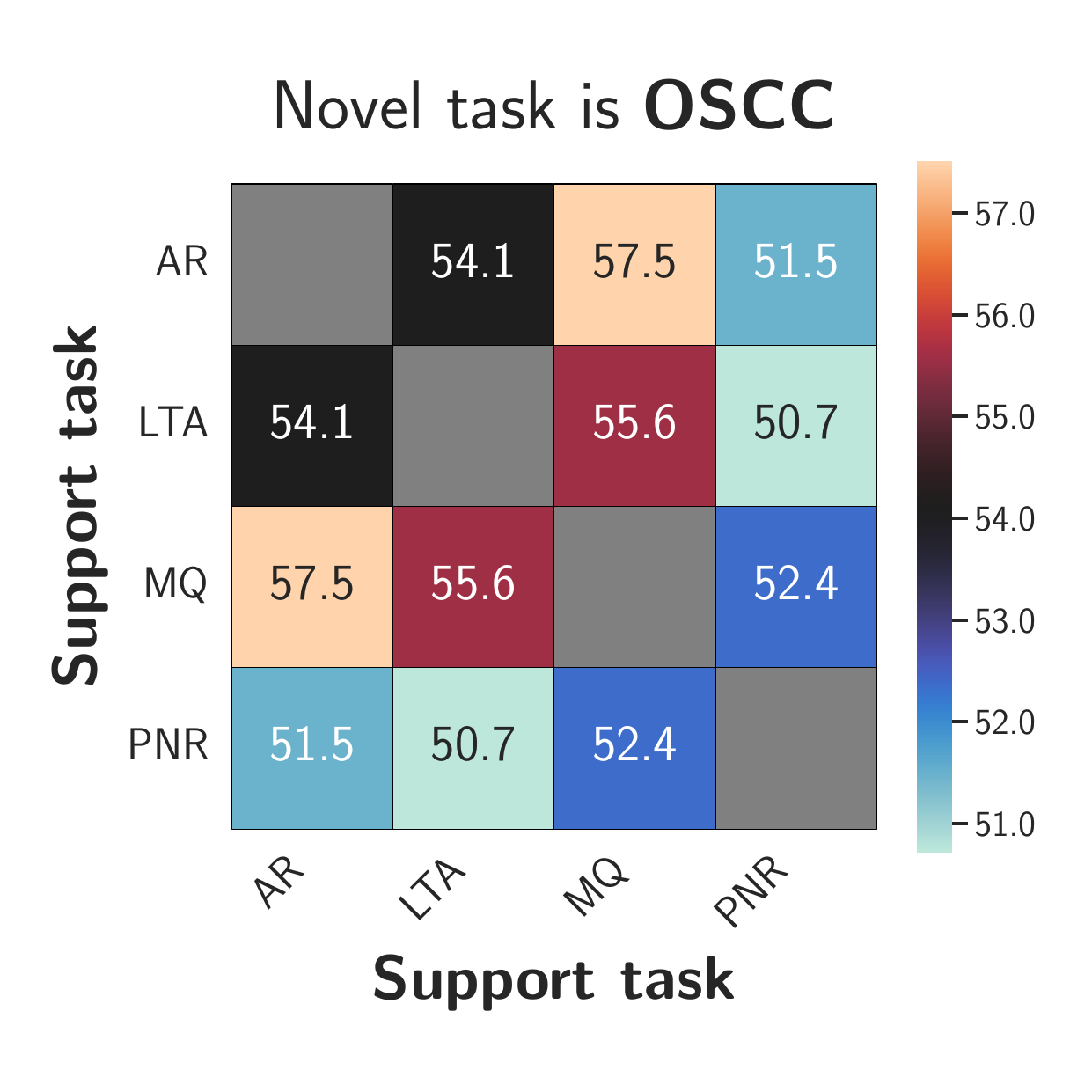}
    \end{minipage}
    \hfill
    \begin{minipage}{0.2425\textwidth}
        \includegraphics[trim=0.5cm 0.5cm 0.5cm 0.5cm,height=4cm]{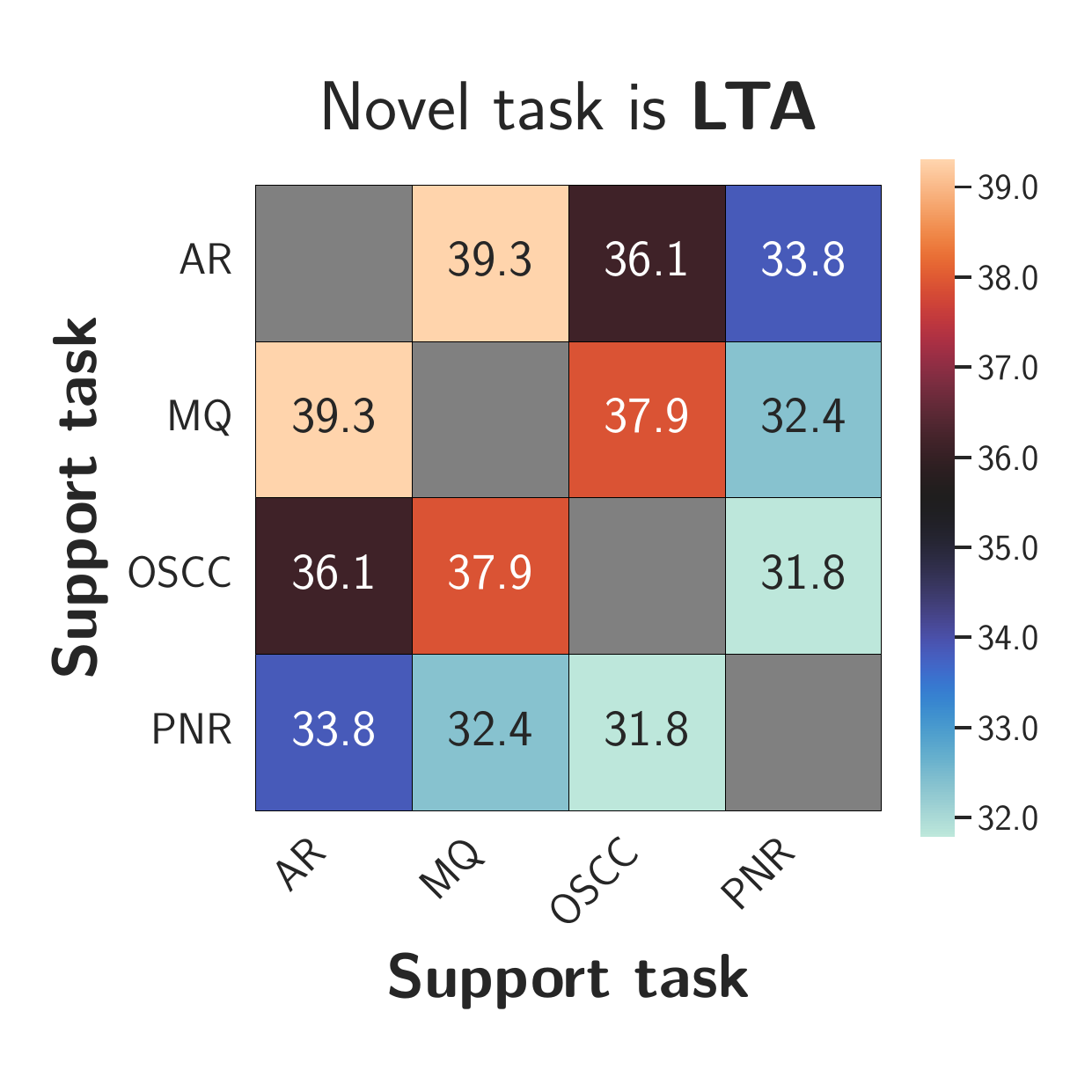}
    \end{minipage}
    \hfill
    \begin{minipage}{0.2425\textwidth}
        \includegraphics[trim=0.5cm 0.5cm 0cm 0.5cm,height=4cm]{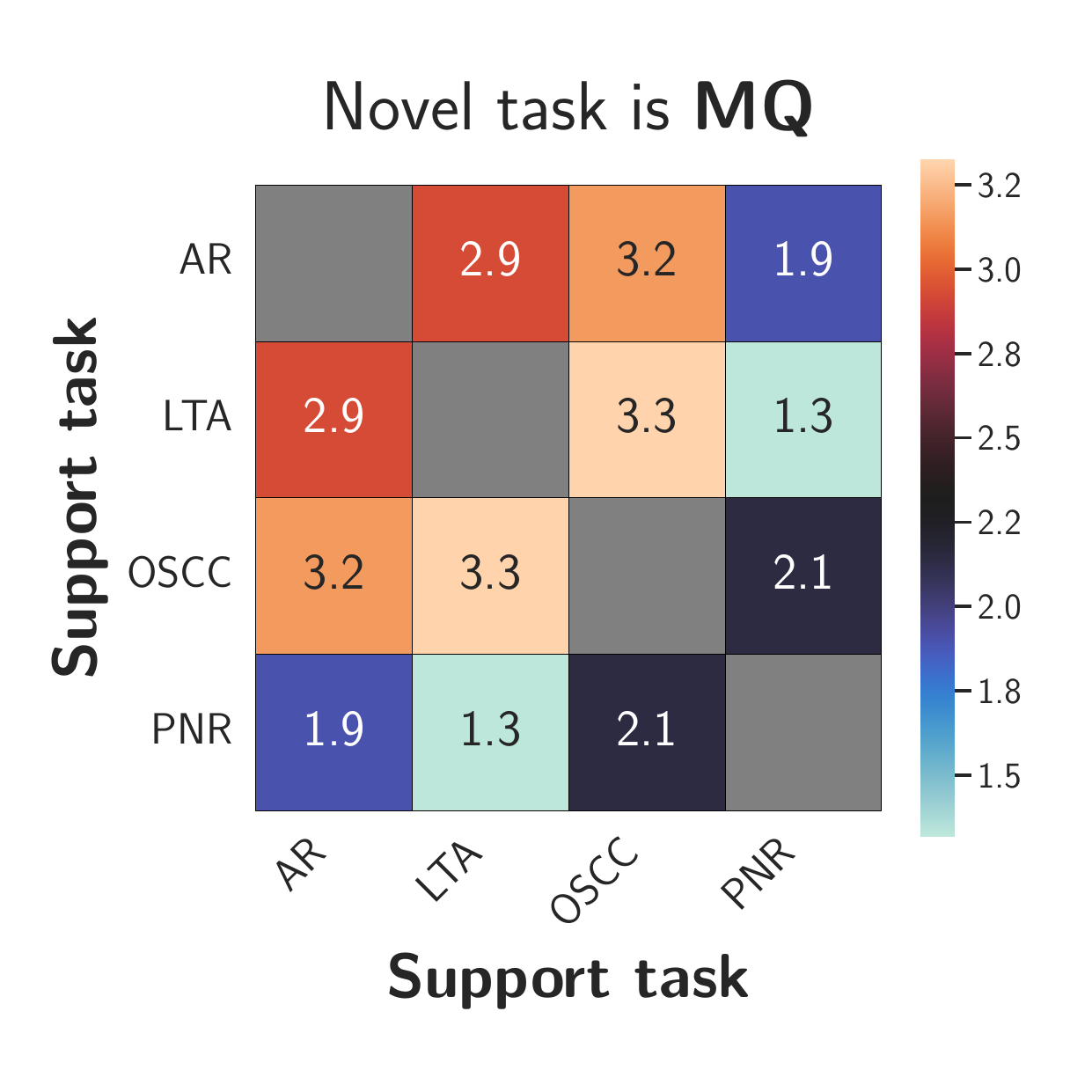}
    \end{minipage}
    \hfill
    \caption{\textbf{Activations consensus for different \emph{novel tasks}.} Activations consensus between two \emph{support tasks} is defined as the percentage of their prototypes corresponding to the same label activated by the two tasks. Fine-grained tasks, \ie, AR, OSCC and LTA, have higher average consensus. On the contrary, MQ has lower average consensus.}
    \label{fig:consensus}
    \vspace{-0.3cm}
\end{figure*}

\subsection{Benchmarks}\label{sec:exp_benchmarks}
We compare \ours on the test set of MQ and LTA benchmarks, to validate the improvements and soundness of our approach.
In this setting, a fair comparison between methods is challenging because of the use of different backbones or feature extractors, supervision levels, ensemble strategies, and challenge-specific tuning, such as training also on the validation set.

\smallskip
\smallskip
\noindent\textbf{Moment Queries (MQ).}
We compare different approaches using EgoVLP features and with the official \egofourd baseline in Table~\ref{tab:test_mq}.
VSGN~\cite{zhao2021video} is a two-stages method featuring a pyramid network to exploit cross-scale correlations in the input video.
ActionFormer~\cite{zhang2022actionformer} is a single-stage method that combines a multi-scale transformer encoder with a lightweight convolutional decoder.
ASL~\cite{shao2023action} extends ActionFormer by reweighting the predictions based on their distance from the corresponding ground truth segment.
ASL is a much larger model in terms of trainable parameters than \ours (350.7 vs. 37.1 MParams) and the test-set results are obtained with an ensemble of three models, each trained with different hyperparameters on the combination of the training and validation splits. We include this model in our analysis because of its relevance and use of EgoVLP features, although it is not directly comparable with the other approaches.
\begin{table}[t]
    \centering
    \caption{Test-set results for Moment Queries (MQ).}\label{tab:test_mq}
    \vspace{-0.295cm}
    \footnotesize
    \setlength{\tabcolsep}{5pt}
    \begin{tabularx}{1.0\columnwidth}{@{}Xc|cccc|c@{}}
        \toprule
                                                            &                   & \multicolumn{4}{c|}{\textbf{Validation mAP @ IoU}} & \textbf{Test mAP}                                                          \\
        \textbf{Method}                                     & \textbf{Features} & \textbf{0.1}                                       & \textbf{0.3}      & \textbf{0.5}     & \textbf{Avg}     & \textbf{Avg}     \\
        \midrule
        \egofourd Baseline~\cite{ego4d}                     & SlowFast          & 9.10                                               & 5.76              & 3.41             & 6.03             & 5.68             \\
        \midrule
        VSGN~\cite{zhao2021video}                           & EgoVLP            & 16.6                                               & 11.5              & 6.57             & 11.4             & 10.3             \\
        ActionFormer$^\dagger$~\cite{zhang2022actionformer} & EgoVLP            & 26.8                                               & 20.6              & 14.5             & 20.6             & 17.5             \\ % train on train
        ASL$^\ddagger$~\cite{shao2023action}                & EgoVLP            & \textbf{29.5}                                      & \textbf{23.0}     & \textbf{16.1}    & \textbf{22.8}    & \textbf{22.3}    \\ % train on train+val
        \textbf{\ours}                                      & EgoVLP            & \underline{27.0}                                   & \underline{21.0}  & \underline{15.2} & \underline{21.0} & \underline{18.0} \\
        \bottomrule
    \end{tabularx}
    \begin{tablenotes}[flushleft]
        \scriptsize
        \item \noindent We report mAP at different thresholds and average mAP in \texttt{[0.1:0.1:0.5]} on the validation and test sets of Moment Queries (MQ). Best results in \textbf{bold}, second best underlined. \noindent $^\dagger$ Reproduced results on the test set (not present in the original paper). \noindent $^\ddagger$ ASL~\cite{shao2023action} is a considerably larger model (350.7~MParams) compared to \ours (37.1~MParams).
        Also, models are trained on both train and validation splits and three different models are ensembled at test-time for better performance~\cite{shao2023action_challenge}.
    \end{tablenotes}
    \vspace{-3mm}
\end{table}

In particular, \ours significantly outperforms VSGN and ActionFormer, despite having a generic architecture not specifically designed for the task.

\smallskip
\smallskip
\noindent\textbf{Long Term Anticipation (LTA).}
We compare different approaches for the LTA task in Table~\ref{tab:test_lta_oscc}.
In particular, we distinguish between \emph{vision-based} and \emph{LLM-based} approaches, with the former relying only on visual reasoning and the latter integrating LLMs into their pipeline.
\ours achieves SOTA performance on the \emph{noun} and \emph{action} metrics in the \emph{vision-based} category, with similar performance compared to \ourscvpr on the \emph{verb} metric.
\begin{table}[t]
    \caption{Test-set results for Long Term Anticipation (LTA).}\label{tab:test_lta_oscc}
    \vspace{-0.25cm}
    \begin{tabularx}{1.0\columnwidth}{@{}Xcccc@{}}
        \toprule
        \textbf{Method}                    & \textbf{Version} & \textbf{Verb ED}  & \textbf{Noun ED}  & \textbf{Action ED} \\
        \midrule

        \rowcolor{lightgray!50} \multicolumn{5}{c}{Vision-based approaches}                                                \\
        \midrule

        SlowFast~\cite{ego4d}              & \textit{v1}      & 0.739             & 0.780             & 0.943              \\

        EgoT2~\cite{egot2}                 & \textit{v1}      & \underline{0.722} & 0.764             & 0.935              \\

        HierVL~\cite{hiervl}               & \textit{v1}      & 0.724             & \underline{0.735} & 0.928              \\

        I-CVAE~\cite{mascaro2023intention} & \textit{v1}      & 0.741             & 0.740             & 0.930              \\

        \ourscvpr~\cite{egopack}           & \textit{v1}      & \textbf{0.721}    & \underline{0.735} & \underline{0.925}  \\

        \textbf{\ours}                     & \textit{v1}      & 0.726             & \textbf{0.716}    & \textbf{0.924}     \\

        \midrule
        \rowcolor{lightgray!50} \multicolumn{5}{c}{LLM-based approaches}                                                   \\
        \midrule

        AntGPT~\cite{zhaoantgpt}           & \textit{v1}      & \underline{0.658} & \underline{0.655} & \underline{0.881}  \\

        PALM~\cite{kim2025palm}            & \textit{v1}      & \textbf{0.656}    & \textbf{0.640}    & \textbf{0.861}     \\

        \bottomrule
    \end{tabularx}
    \begin{tablenotes}[flushleft]
        \scriptsize
        \item \noindent We report Verb, Noun and Action Edit Distance on the test set of Long Term Anticipation (LTA), separately for \emph{vision-based} and \emph{LLM-based} approaches.
    \end{tablenotes}
    \vspace{-0.3cm}
\end{table}

\subsection{Qualitative results}
\label{sec:exp_qualitative}
In this section, we analyze how \ours leverages knowledge abstractions from the \emph{support tasks} (collected in the form of prototypes) to aid the learning of a \emph{novel task}.
Specifically, we visualize the \emph{activated prototypes} (\ie the set of prototypes each \emph{support task} looks at) during the interaction process of \ours across different novel tasks and quantify task activation consensus, a measure of the complementarity among support tasks in aiding the learning of a novel task.

\smallskip
\smallskip
\noindent\textbf{Prototypes activations.}
We show in Fig.~\ref{fig:activations} the activation frequency for the task-specific prototypes for a subset of \emph{novel tasks}, considering the Top-20 most activated prototypes.
Due to the large number of prototypes, we aggregate them based on their verb labels to enhance the readability of the plots.
Some tasks, \ie OSCC and LTA, also show more similar activations frequencies for the prototypes corresponding to the same label while Moment Queries have a much larger variability in prototypes activations.

\smallskip
\smallskip
\noindent\textbf{Activations consensus.}
The goal of this analysis is to showcase how a \emph{novel task} can leverage the perspectives from a set of \emph{support tasks}, reusing previously learned knowledge stored in the form of prototypes.
To this end, we expect \ours to extract complementary cues from each \emph{support task}.
We define the \emph{activations consensus} as the degree to which different tasks activate prototypes corresponding to the same label for a given sample of the \emph{novel task}.
A low consensus suggests that the support tasks capture more diverse cues, \ie different tasks activate different prototypes, whereas a high consensus indicates that activations are more coherent across tasks.
We show in Fig.~\ref{fig:consensus} the average activation consensus for different novel tasks.
Fine-grained tasks, \eg AR, LTA and OSCC, have higher average consensus compared to MQ.
We attribute this difference to the implementation of the interaction process for these two groups of tasks.
In fine-grained tasks, the interaction process is applied on the sample-level aligned features. On the contrary, for MQ the interaction is applied to node-level features, without any alignment due to the nature of the task, as previously stated in Sec.~\ref{ts-design}.
Therefore, a substantially higher number of nodes per video interact with the task-specific prototypes.
These nodes may correspond to background regions of the video or to segments of an activity that are insufficiently discriminating.
The low average activations consensus (Fig.~\ref{fig:consensus}) and high diversity in prototypes' activations across tasks (Fig.~\ref{fig:activations}) show how \ours is effectively integrating different perspectives for the Moment Queries task.

\smallskip
\smallskip
\noindent\textbf{Activation frequency.}
\begin{figure*}[tbp]
    \centering
    \includegraphics[trim=0cm 0cm 0cm 0cm,height=9cm]{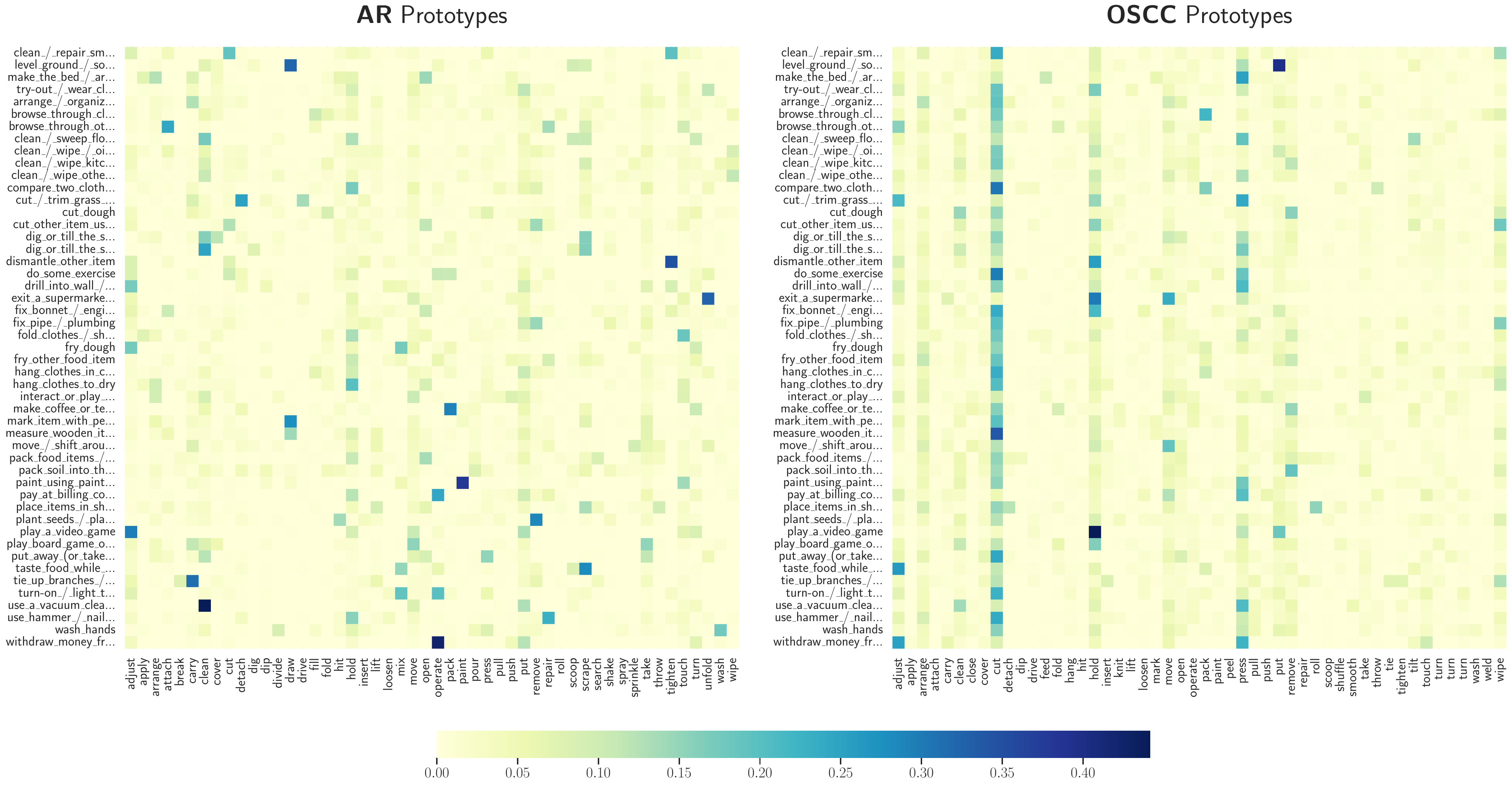}
    \caption{\textbf{Activation frequency of the prototypes from the \emph{support tasks} when the \emph{novel task} is Moment Queries (MQ).}
        For each task from the MTL pre-training phase, we plot the distribution of closest prototypes in the \ours interaction phase. For readability, we restrict our analysis to the top 50 most predicted labels and activated prototypes. \textit{Best viewed on a screen.}}\label{fig:mq_activations}
    \vspace{-0.3cm}
\end{figure*}
We show in Fig.~\ref{fig:mq_activations} the most activated prototypes for different \emph{support tasks} when the \emph{novel task} is MQ.
To enhance readability, we select the 50 most predicted labels and 50 most activated prototypes.
Overall, we observe that the activations of the AR task are quite sparse, indicating that the novel task looks at very different perspectives from these tasks.
On the contrary, the activations of the OSCC task are more uniform across different MQ labels. This is because these tasks focus on detecting object state changes in the video, which are typically associated with a subset of specific actions, such as \emph{cut} or \emph{mix}.
As a result, only a subset of prototypes from these \emph{support tasks} is actually activated by the novel task, as can be seen from the stripes in the plots.

\begin{figure}[t]
    \hfill
    \begin{minipage}{0.485\columnwidth}
        \includegraphics[trim=0.5cm 0.5cm 0cm 0.5cm,height=4cm]{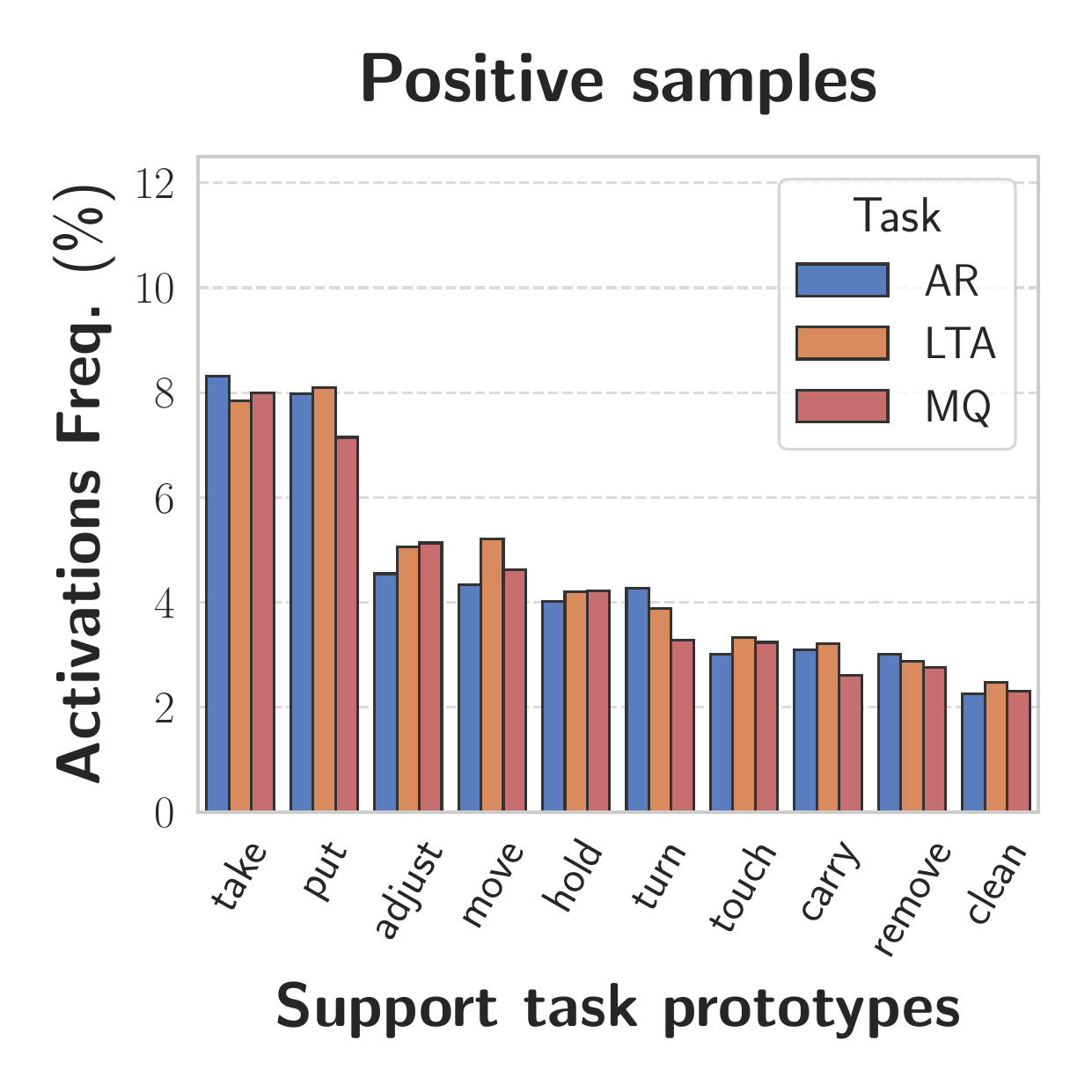}
    \end{minipage}
    \hfill
    \begin{minipage}{0.485\columnwidth}
        \includegraphics[trim=0cm 0.5cm 0.5cm 0.5cm,height=4cm]{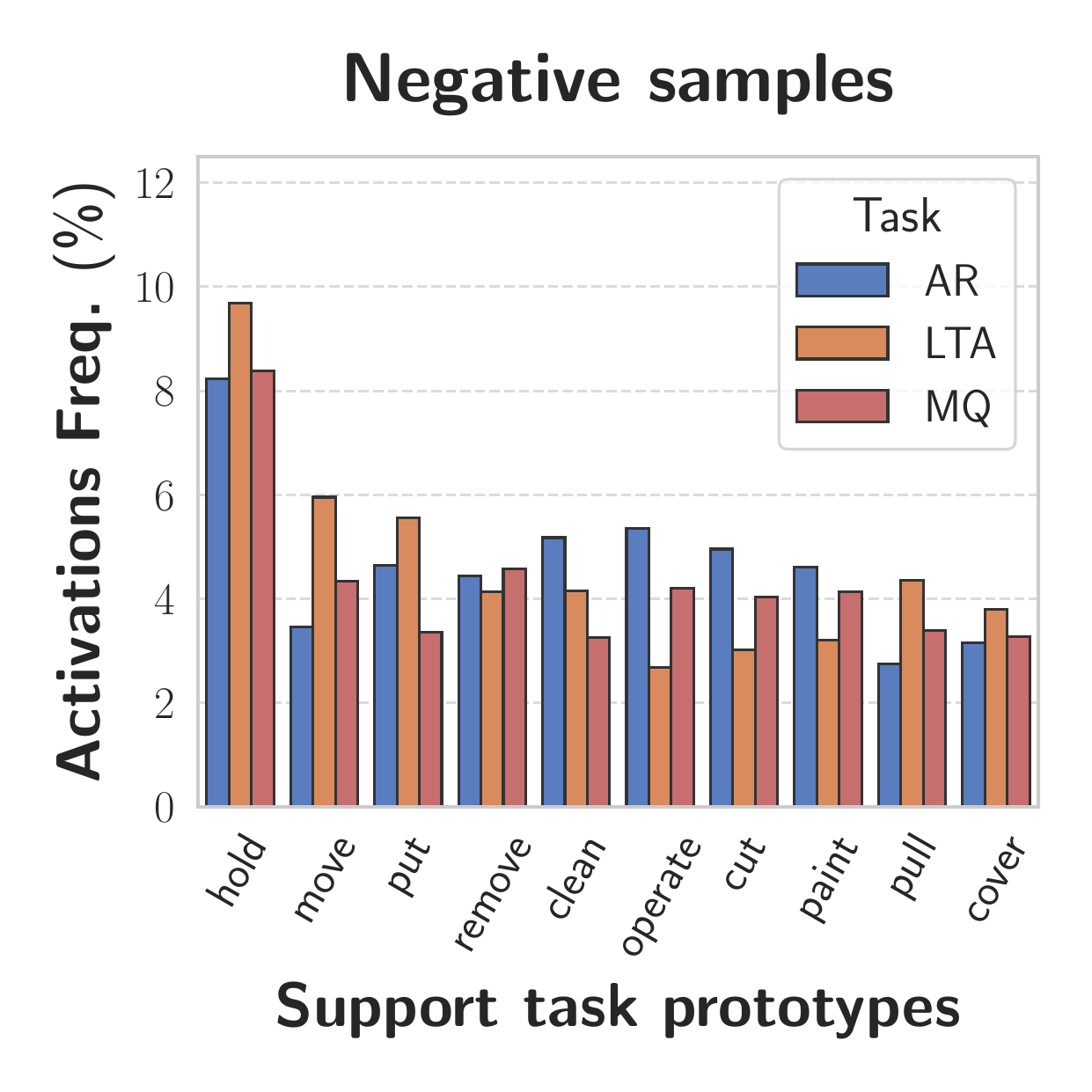}
    \end{minipage}
    \hfill
    \caption{\textbf{Activation frequency of the prototypes from the \emph{support tasks} when the \emph{novel task} is OSCC}, separately for the positive and negative correct predictions. Positive samples tend to focus more on prototypes whose verb could be associated with an object state change, \eg, \emph{take} or \emph{put}, compared to negative samples.
    }
    \label{fig:oscc_activations}
    \vspace{-0.3cm}
\end{figure}
Similarly, we show in Fig.~\ref{fig:oscc_activations} the most activated prototypes when the \emph{novel task} is OSCC.
We consider separately correctly predicted segments that contain an object state change (\emph{positive}) or not (\emph{negative}).
Positive samples tend to focus more on prototypes whose verb could be associated with an object state change, \eg \emph{take} or \emph{put}, compared to negative samples.

\section{Conclusions}\label{sec:conclusions}
We presented \ours, an extension of \ourscvpr that enables knowledge sharing between egocentric vision tasks with different temporal granularity.
\ours is built on a unified temporal architecture that progressively learns more abstract representations of the input video, using a novel GNN layer specifically designed to incorporate strong temporal reasoning.
We evaluate our approach in a \emph{novel task learning} setting, in which a model is first trained on set of known \emph{support tasks} and then has to leverage the knowledge obtained from such tasks to improve the learning process of a \emph{novel task}.
We validate \ours on five \egofourd tasks, covering a wide range of temporal granularities, from sub-second actions to long-range activities. Results show the effectiveness of our approach in knowledge reuse, outperforming single-task and multiple-task baselines, as well as task translation approaches that seek to share knowledge across tasks but lack explicit knowledge abstraction.
Our work emphasizes the importance of prior knowledge and task perspectives in learning novel tasks, focusing on how task-specific knowledge is represented and utilized. Furthermore, through our proposed unified video understanding architecture, we demonstrate that leveraging diverse task perspectives in egocentric vision, even across varying temporal granularities, leads to more comprehensive and human-like video understanding.

\section*{Acknowledgments}
This study was carried out within the FAIR - Future Artificial Intelligence Research and received funding from the European Union Next-GenerationEU (PIANO NAZIONALE DI RIPRESA E RESILIENZA (PNRR) – MISSIONE 4 COMPONENTE 2, INVESTIMENTO 1.3 – D.D. 1555 11/10/2022, PE00000013). This manuscript reflects only the authors’ views and opinions, neither the European Union nor the European Commission can be considered responsible for them. We acknowledge the CINECA award under the ISCRA initiative, for the availability of high performance computing resources and support. Antonio Alliegro and Tatiana Tommasi also acknowledge the EU project ELSA - European Lighthouse on Secure and Safe AI (grant number 101070617).

\bibliographystyle{IEEEtran}
\bibliography{main}

\vspace{-33pt}
\begin{IEEEbiography}[{\includegraphics[width=1in,height=1.25in,clip,keepaspectratio]{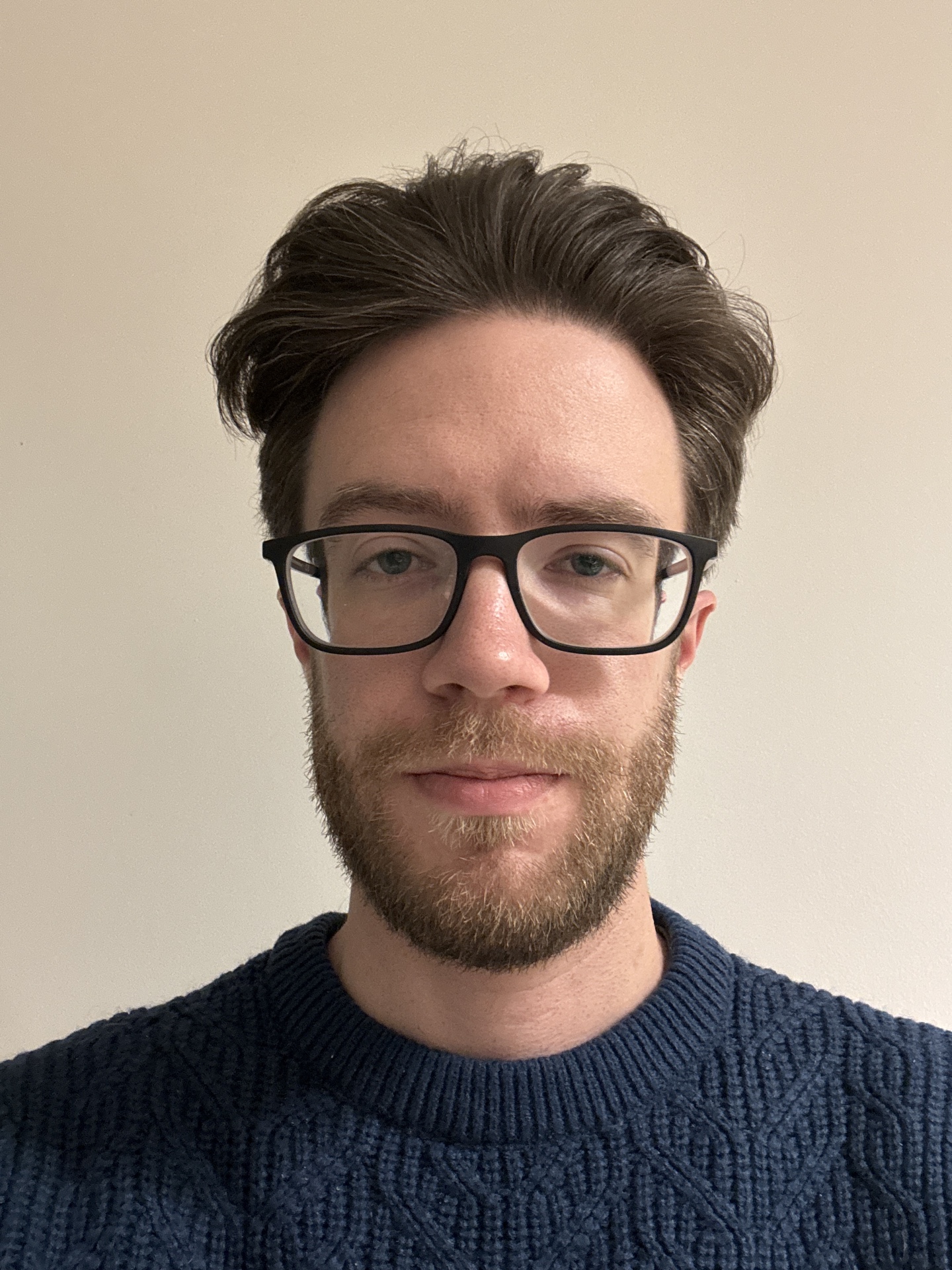}}]{Simone Alberto Peirone}
received the B.Sc. and M.Sc. degrees in Computer Engineering from the Polytechnic University of Turin, in 2020 and 2022 respectively. He is currently pursuing the Ph.D. degree with the Visual and Multimodal Applied Learning Laboratory, Turin, working under the supervision of Prof. Giuseppe Averta. His research focuses on egocentric vision and applications of graph neural networks to video understanding.
\end{IEEEbiography}

\vspace{-33pt}
\begin{IEEEbiography}[{\includegraphics[width=1in,height=1.25in,clip,keepaspectratio]{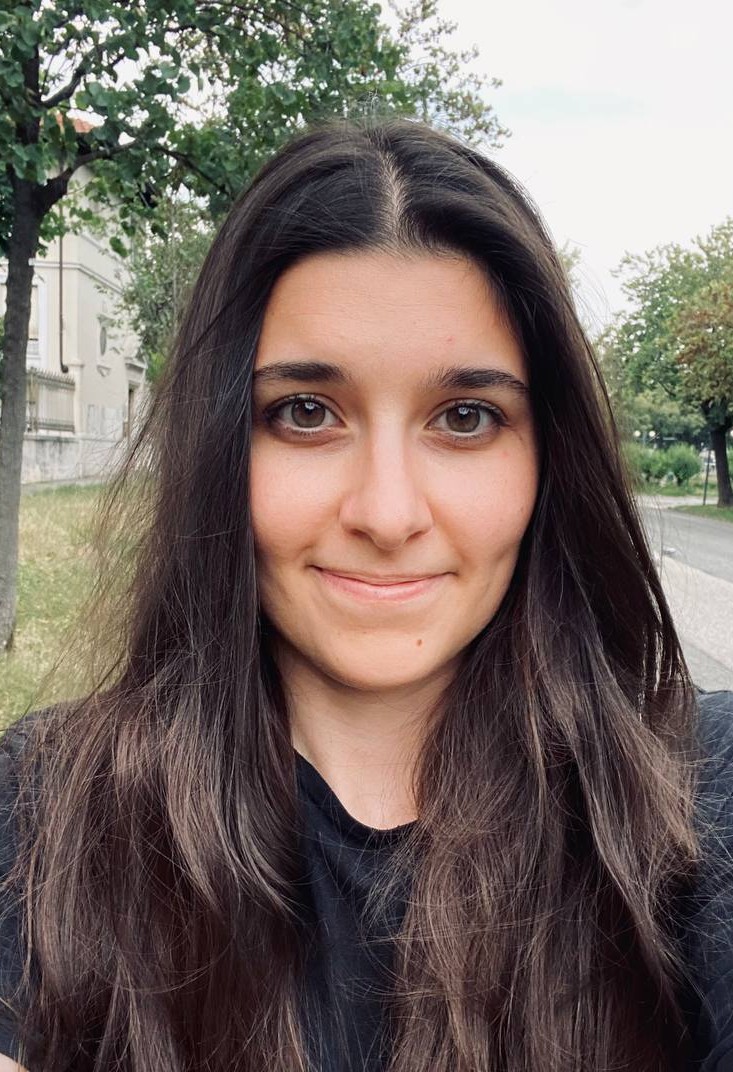}}]{Francesca Pistilli}
received the M.Sc. degree in electronic engineering from the Polytechnic of Turin, in 2019, the M.Sc. degree in electrical and computer engineering from the University of Illinois at Chicago, Chicago, IL, USA, in 2020, and the Ph.D. degree from the Image Processing and Learning Group (IPL), Polytechnic of Turin, in 2023. She is currently Assistant Professor at the Polytechnic of Turin. Her current research interests lie at the intersection between computer vision and robotics.
\end{IEEEbiography}

\vspace{-33pt}
\begin{IEEEbiography}[{\includegraphics[width=1in,height=1.25in,clip,keepaspectratio]{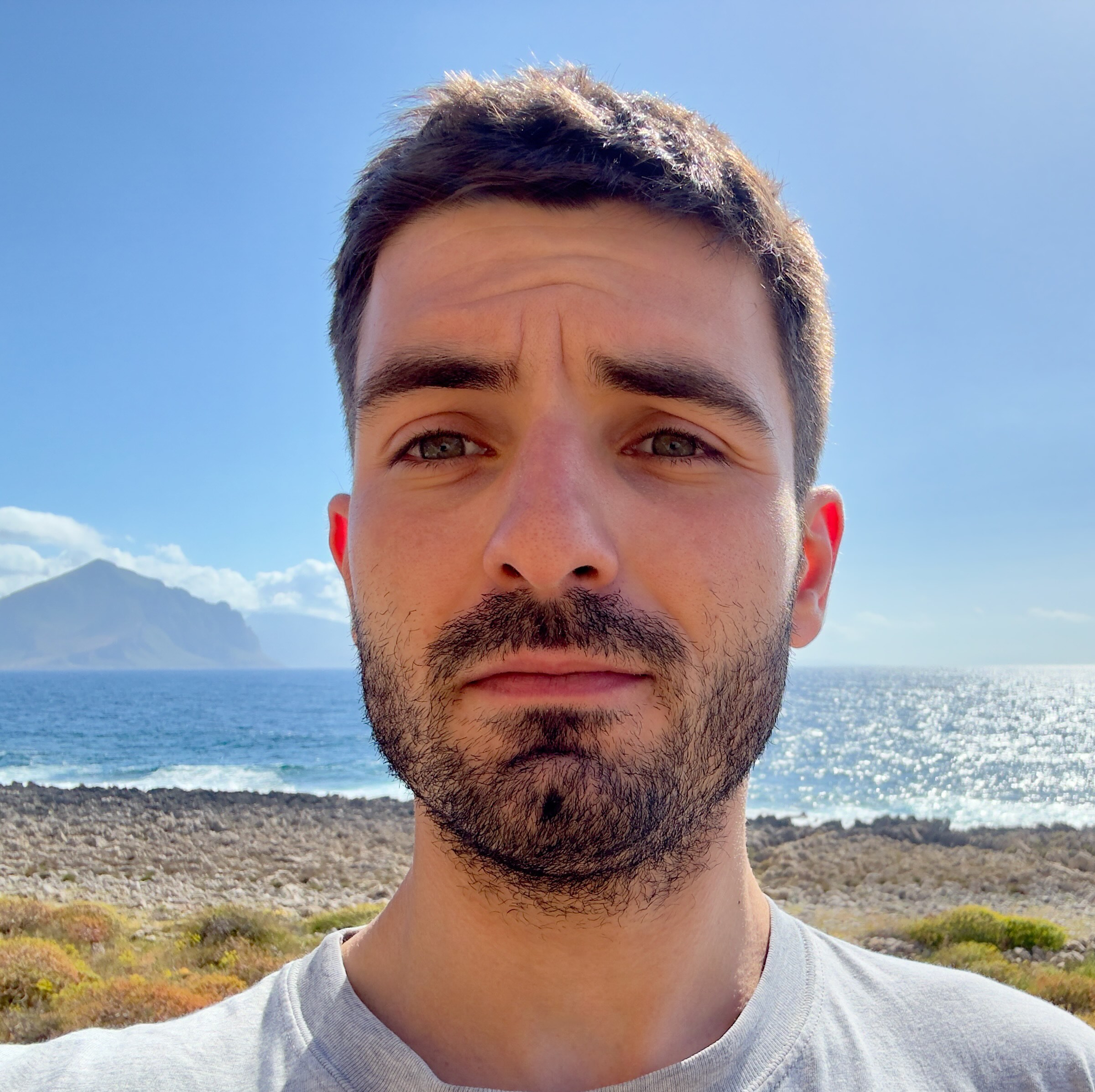}}]{Antonio Alliegro}
is currently a Postdoctoral Researcher with the Polytechnic University of Turin. His research focuses on 3D understanding and its application to real-world scenarios, including research on reducing the synth-to-real domain gap and open-set 3D recognition. 
He has published multiple papers presented at prestigious computer vision conferences and journals such as CVPR, IROS, NeurIPS, and RA-L. 
Additionally, he has contributed as a reviewer at various academic events.
\end{IEEEbiography}

\begin{IEEEbiography}[{\includegraphics[width=1in,height=1.25in,clip,keepaspectratio]{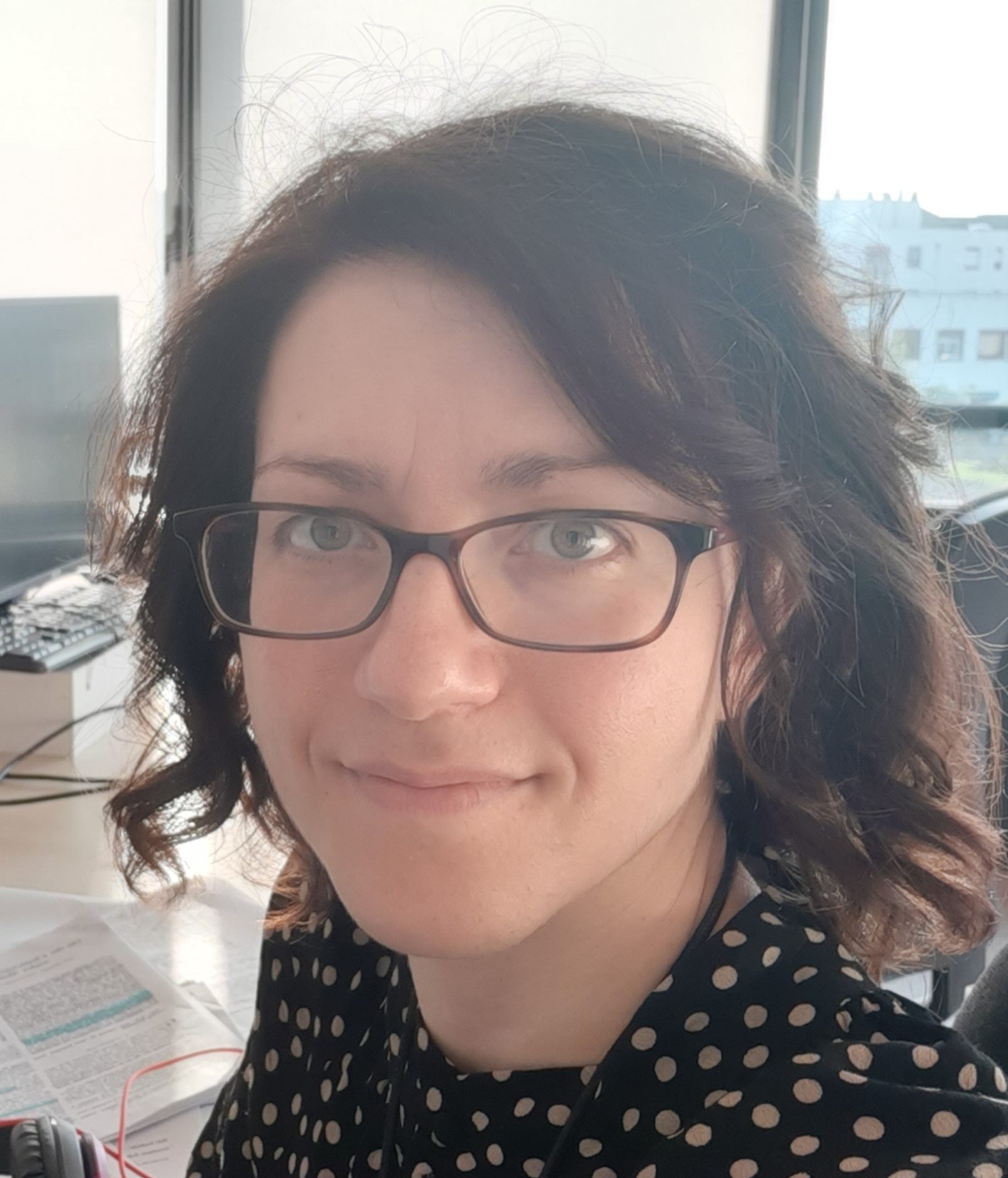}}]{Tatiana Tommasi}
received the Ph.D. degree from EPFL Lausanne, in 2013. Subsequently, she undertook a postdoctoral roles in both Belgium and the USA. She holds the position of an Associate Professor with the Control and Computer Engineering Department, Polytechnic University of Turin, and the Director of the ELLIS Unit, Turin. Before her current position, she was an Assistant Professor with Sapienza University in Rome, Italy. Her publication record boasts over 50 papers in top conferences and journals, specializing in machine learning and computer vision. Her expertise lies in the development of theoretically grounded algorithms for automatic learning from images, particularly within the realms of robotics, medical applications, and human–machine interaction. She pioneered the field of transfer learning in computer vision and possesses extensive experience in areas, such as domain adaptation, generalization, multimodal learning, and open-set learning. She is an Associate Editor of IEEE Transaction on Pattern Analysis and Machine Intelligence and IEEE Transaction on Emerging Topics in Computing.
\end{IEEEbiography}

\begin{IEEEbiography}[{\includegraphics[width=1in,height=1.25in,clip,keepaspectratio]{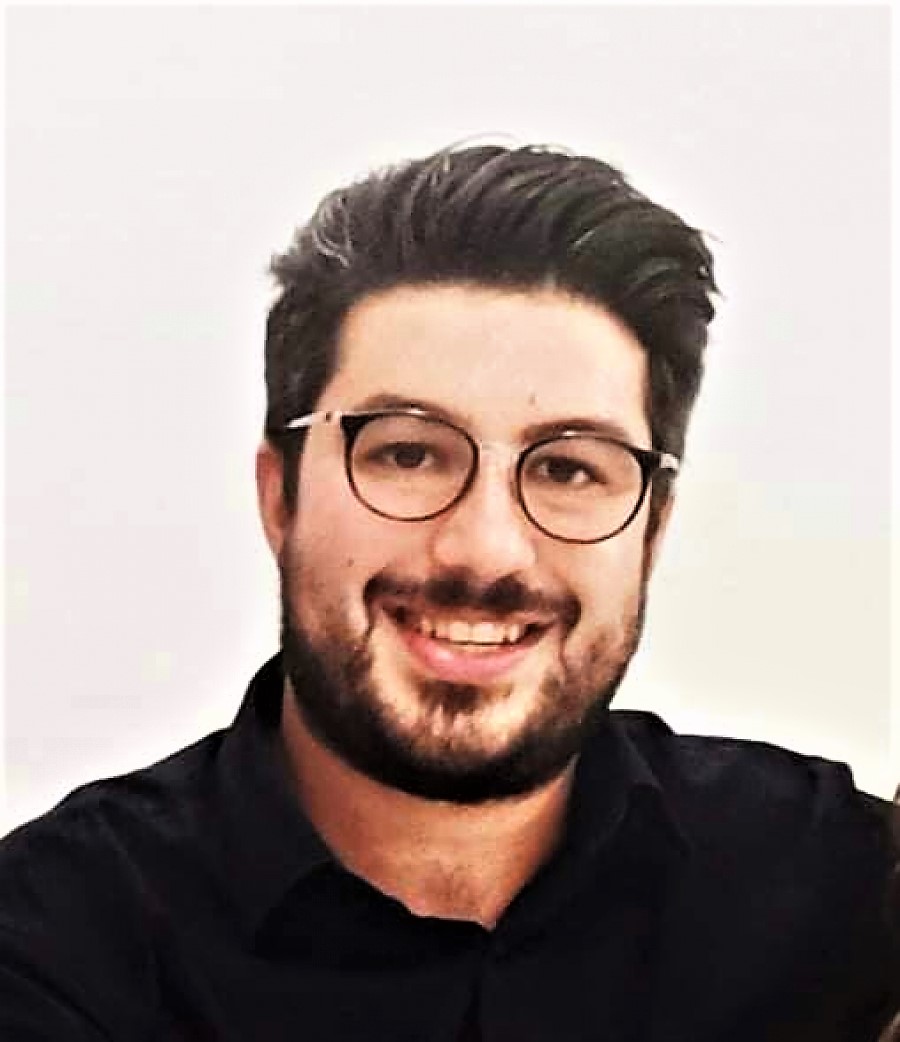}}]{Giuseppe Averta}
received the PhD in robotics from the University of Pisa in 2020. In 2019, he was a Visiting Student with the Eric P. and Evelyn E. Newman Laboratory, Biomechanics and Human Rehabilitation Group, MIT. He is currently an Assistant Professor of robotics and machine learning with the Polytechnic of Turin. He is also an Italian Institute of Technology Alumnus. His current research interests include the development of a truly embodied intelligence for human-robot cooperation, with research activities in human action recognition, deep learning for egocentric vision, human-inspired design, planning, and control guidelines for autonomous, collaborative, assistive, and prosthetic robots.
\end{IEEEbiography}

\vfill

\end{document}